%% file: main.tex
\documentclass[10pt,twocolumn,letterpaper]{article}

\usepackage[pagenumbers]{cvpr}

\input{preamble}
\usepackage[accsupp]{axessibility}

\definecolor{cvprblue}{rgb}{0.21,0.49,0.74}
\usepackage[pagebackref,breaklinks,colorlinks,allcolors=cvprblue]{hyperref}

\def\paperID{6098}
\def\confName{CVPR}
\def\confYear{2026}

\title{TrafficAlign: Aligning Large Language Models for Traffic Scenario Generation}

\author{Zhi Tu\qquad
Liangkun Niu\qquad
Tianyi Zhang\\
Purdue University\\
{\tt\small \{tu85, niu61, tianyi\}@purdue.edu}\\
{\small \url{https://github.com/TrafficComposer/TrafficAlign}}
}

\begin{document}
\maketitle
\input{sec/00_abstract}    
\input{sec/01_intro}
\input{sec/02_related}

\input{sec/03_method}
\input{sec/04_experiments}

\input{sec/05_limitations}
\input{sec/06_conclusion}
\input{sec/07_acknowledgement}
{
    \small
    \bibliographystyle{ieeenat_fullname}
    \bibliography{sec/00_references}
}

\clearpage
\renewcommand{\maketitlesupplementary}{%
  \newpage
  \onecolumn
  {\centering
   \Large \textbf{\thetitle}\\
   \vspace{0.5em}Supplementary Material \\[1.0em]}%
}
\input{sec/X_suppl}

\end{document}

%% file: preamble.tex
\usepackage{microtype}

\renewcommand{\paragraph}[1]{\vspace{.5em}\noindent\textbf{#1}}

\usepackage{soul}
\setuldepth{foobar}

\usepackage{graphicx}	
\usepackage{amsmath}	
\usepackage{amssymb}	
\usepackage{booktabs}
\usepackage{times}
\usepackage{epsfig}
\usepackage{caption}
\usepackage{float}
\usepackage{placeins}
\usepackage{color, colortbl}
\usepackage{stfloats}
\usepackage{enumitem}
\usepackage{tabularx}
\usepackage{xstring}
\usepackage{multirow}
\usepackage{xspace}
\usepackage{url}
\usepackage{xcolor}
\usepackage[hang,flushmargin]{footmisc}

\newcommand{\supp}{Supplementary Material}

\newcommand{\tool}{\textsc{TrafficAlign}}

\definecolor{myresultgreen}{HTML}{D5E5CF}

\newcommand{\numads}{three}
\newcommand{\numdataset}{six}
\newcommand{\numbaselines}{five}
\newcommand{\numllmbaseline}{five}

\newcommand{\crimprove}{10.8\%}
\newcommand{\osimprove}{35.6\%}

\newcommand{\ftcrimprovefromraw}{36.1\%}

\newcommand{\result}[1]{#1}

\usepackage{tcolorbox}
\tcbuselibrary{skins, breakable}
\usepackage{fvextra}
\definecolor{slategray}{RGB}{112,128,144}

%% file: sec/00_abstract.tex
\begin{abstract}
Recent research has investigated the use of large language models (LLMs) to generate traffic scenarios for autonomous driving.
However, pretrained LLMs often fail to align with real-world traffic distributions. In this work, we present {\tool}, an automated framework that synthesizes traffic scenarios based on real-world driving videos, performs data validation, and aligns LLMs with the synthesized scenarios.
The evaluation shows that traffic scenarios generated by {\tool} are highly effective, revealing up to {\crimprove} more collisions on average across three autonomous driving models than state-of-the-art methods.
Furthermore, fine-tuning these driving models with {\tool}-generated scenarios significantly reduced collision rates by {\ftcrimprovefromraw} compared with the original models.
A qualitative study using traffic datasets from {\numdataset} geographically diverse regions shows that {\tool}-generated scenarios exhibit strong alignment with corresponding traffic distributions in these regions. 
\end{abstract}

%% file: sec/01_intro.tex
\section{Introduction}
\label{sec:intro}
Despite significant progress in autonomous driving, traffic accidents caused by autonomous vehicles continue to occur~\cite{nhtsa2025standing}. 
This highlights the need for comprehensively testing driving models.
In recent years, simulation-based testing has emerged as a cost-effective and scalable alternative method to real-world road tests. However, a critical challenge in simulation-based testing lies in effectively generating realistic traffic scenarios.

Recent studies have explored the use of large language models (LLMs) to automatically generate traffic scenarios~\cite{elmaaroufi2024scenicnl, sheng2025talk2traffic, DBLP:conf/aaai/Aiersilan25, DBLP:conf/corl/TanIW0K23, trafficcomposer, DBLP:journals/tse/DengTYZZZ25, tang2024legend}.
However, most of these approaches require humans to write natural language descriptions of desired traffic scenarios in order to generate the simulation scripts. As a result, this manual input could become labor-intensive in large-scale testing.
One exception is ChatScene~\cite{Zhang_2024_CVPR}, which reduces the dependency on manual effort by directly prompting LLMs to generate natural language descriptions of traffic scenarios.
However, LLMs lack sufficient traffic domain knowledge and also tend to generate homogeneous scenarios.
As a result, they struggle to generate scenarios that align with real-world traffic characteristics and distributions, especially for specific geographic contexts such as large cities, small towns, and mountainous regions.

To address these challenges, we propose {\tool}, an automatic framework that synthesizes traffic scenarios from real-world driving videos, checks data validity, and aligns LLMs with the synthesized scenarios. 
To effectively validate the synthesized  scenarios, {\tool} converts the natural language description of each scenario into a formal representation in a domain-specific language (DSL).
Compared with natural language, this DSL representation enables systematic analysis and rigorous checking. The identified issues are then fed back to the data synthesizer for refinement or regeneration. 
Finally, {\tool} uses the validated scenarios to fine-tune LLMs so that it can generate realistic scenarios that align with the real-world traffic distributions of a target region.

We conduct a comprehensive evaluation by comparing {\tool} with ChatScene~\cite{Zhang_2024_CVPR}, two adversary-based baselines~\cite{ding2020learning, wang2021advsim}, two rule-based baselines~\cite{carla2019scenariorunner, zhang2022adversarial}, and {\numllmbaseline} LLM baselines.
We first evaluate the testing effectiveness of traffic scenarios generated by each method on {\numads} driving models using the SafeBench platform~\cite{10.5555/3600270.3602131}.
{\tool} reveals 
up to {\crimprove} more collisions across the three driving models than the strongest baseline (Section~\ref{sec:exp_adstest}). 
We further demonstrate that these generated scenarios can be used to improve the safety of driving models. 
Specifically, the driving models fine-tuned with {\tool}-generated scenarios 
reduce collision rate by {\ftcrimprovefromraw} compared with the original models before fine-tuning when evaluated on a held-out set (Section~\ref{sec:exp_ft}). 
In addition, a qualitative study across geographically diverse regions shows that scenarios generated by {\tool} align more closely with the real-world traffic distributions (Section~\ref{sec:exp_vis}).

Our main contributions are concluded as follows:
\begin{itemize}[leftmargin=*, labelindent=\parindent]
    \item We propose {\tool}, an automated framework that synthesizes traffic scenarios from real-world driving videos, validates their quality, and aligns LLMs with these scenarios.
    \item {\tool}-generated scenarios show high effectiveness in testing {\numads} autonomous driving models, revealing a collision rate up to {\crimprove} higher than the state-of-the-art baseline, ChatScene~\cite{Zhang_2024_CVPR}.
    \item Fine-tuning the driving models with {\tool}-generated scenarios reduces the average collision rate by up to {\ftcrimprovefromraw} compared with the original models, indicating {\tool}'s effectiveness in improving driving model performance.
\end{itemize}

%% file: sec/02_related.tex
\section{Related Work}
\label{sec:related}

\paragraph{Traffic scenario generation.}
There is a large body of literature on traffic scenario generation. Prior work can be broadly classified into three categories. 
First, \emph{search-based approaches} start from seed scenarios and explore the scenario search space through scenario mutation guided by adversarial learning~\cite{8793740, 7311450, pmlr-v229-zhang23g} or heuristic functions~\cite{li2020av, luo2021targeting, cheng2023behavexplor, abdessalem2018testing, sun2022lawbreaker, kim2022drivefuzz}.
While effective in generating safety-critical scenarios, these approaches often incur a high computational cost and suffer from limited scenario diversity.
Second, \emph{scenario reconstruction approaches}~\cite{9197145, SCANLON2021106454, Yang_2020_CVPR, van2015automated, 10.1145/3338906.3338942, trafficcomposer} generate traffic scenarios from real-world driving data, such as sensor logs and crash reports.
These approaches generate realistic scenarios but suffer from data scarcity.
Third, \emph{rule-based approaches}~\cite{8500632, 9294629, 8317919, tian2022generating} encode traffic laws and physical constraints to compose scenarios. 
While they provide systematic and interpretable approaches for scenario generation, they require extensive manual rule encoding, which significantly limits their scalability.
Moreover, since handcrafted rule sets rarely capture the full spectrum of traffic scenarios, their coverage of safety-critical cases is limited.

Another line of research learns a generative model of traffic scenarios~\cite{Kar_2019_ICCV, devaranjan2020meta, rempe2022generating, chitta2024sledge, NEURIPS2023_d95cb79a, Suo_2021_CVPR}.
Meta-Sim~\cite{Kar_2019_ICCV} learns a probabilistic grammar and optimizes its parameters through differentiable rendering to minimize the domain gap between synthetic and real data, enabling adaptive generation of realistic training scenarios. 
STRIVE~\cite{rempe2022generating} learns a latent generative model of traffic scenarios and performs adversarial optimization to create realistic yet collision-prone scenarios for stress-testing planners. 
SLEDGE~\cite{chitta2024sledge} leverages diffusion transformers to generate scalable, map-conditioned driving environments that capture diverse traffic configurations with controllable density. 
Scenario Diffusion~\cite{NEURIPS2023_d95cb79a} adapts diffusion models for controllable, structured driving scenario generation with constraints such as collision likelihood or traffic flow density. 
These methods model scenarios at the structured bird's-eye-view or trajectory level (e.g., lane graphs, boxes, trajectories), whereas {\tool} evaluates scenarios as 3D traffic simulations in CARLA. 
We therefore do not include them as baselines in our evaluation.

Recent studies have explored the use of LLMs for traffic scenario generation~\cite{DBLP:journals/tse/DengTYZZZ25, trafficcomposer, guo2024sovar, elmaaroufi2024scenicnl, Zhang_2024_CVPR, DBLP:conf/corl/TanIW0K23}.
TARGET~\cite{DBLP:journals/tse/DengTYZZZ25} uses an LLM (GPT-3.5) to generate traffic scenarios from traffic rules in driving handbooks.
However, the generation of simulation scripts relies on human-curated rule-based algorithms.
ScenicNL~\cite{elmaaroufi2024scenicnl} addresses this challenge by combining a comprehensive prompting strategy with a compiler-in-the-loop to generate Scenic~\cite{DBLP:journals/ml/FremontKDGYSS23} scripts from crash reports using LLMs.
However, it depends on human-written crash reports, which are labor-intensive and hard to scale up for massive traffic scenario generation.
ChatScene~\cite{Zhang_2024_CVPR} addresses this limitation by leveraging an LLM's internal knowledge and directly prompting it to design safety-critical scenarios, thereby reducing dependence on manually crafted scenario descriptions.
However, LLMs' generalist knowledge lacks fine-grained, location-specific traffic understanding, causing the generated scenarios to deviate from real-world traffic distributions. 
{\tool} addresses this issue by aligning an LLM's general knowledge with real-world observations.
{\tool} automatically synthesizes scenarios from large-scale real-world driving videos and aligns the LLM with the representations, thereby generating traffic scenarios consistent with real-world traffic distributions.

\paragraph{LLM alignment.}
While there are various methods for LLM alignment, the most related to us are methods that generate high-quality synthetic data for alignment~\cite{wei2024selfcodealign, wang2024codeclm, DBLP:conf/icml/YuanPCLSXW24, DBLP:conf/fat/HuangSLLDTG24, NEURIPS2024_be2e1b68, lin2024flame, DBLP:conf/acl/WangWXTZWCYZLNM25, wang2023self, DBLP:conf/iclr/XuSZG0FTLJ24}.
Self-Instruct~\cite{wang2023self} proposes a semi-automated data synthesis pipeline that prompts a pretrained LLM to create diverse instruction–input–output triples and then applies heuristic validity and duplication filtering to generate high-quality and diverse data for LLM instruction-tuning.
Evol-Instruct~\cite{DBLP:conf/iclr/XuSZG0FTLJ24} introduces an evolutionary data synthesis method that iteratively rewrites instructions from a seed instruction pool into more diverse and complex instructions with failure filtering.
CodecLM~\cite{wang2024codeclm} proposes an LLM-guided instruction generator that maps seed tasks into metadata and decodes them into diverse instruction examples for fine-tuning.
FLAME~\cite{lin2024flame} proposes a factuality-aware data curation pipeline that classifies prompts by factual requirements and uses the LLM itself to generate factual responses, constructing hallucination-resistant fine-tuning data.
In a similar spirit, {\tool} automatically extracts traffic scenarios from videos and uses the synthesized scenario descriptions to align LLMs.
In addition, {\tool} introduces an effective validation method to analyze the semantics of the synthesized data (Section~\ref{sec:method-data-validation}), improving the data quality for LLM alignment.

%% file: sec/03_method.tex
\section{Method}
\label{sec:method}

\input{figs/fig_pipeline}
In this work, we introduce {\tool}, an automated framework that synthesizes traffic scenario representations from real-world driving videos, checks data validity, and aligns LLMs with these representations.
Figure~\ref{fig:pipeline} shows {\tool}'s data synthesis pipeline.
{\tool} first synthesizes traffic scenario descriptions from driving videos using a multimodal LLM (Section~\ref{sec:method-extract}).
To validate data quality, {\tool} translates synthesized scenario descriptions into a formal representation in a domain-specific language (DSL) and performs systematic checks to detect semantically incomplete or irrelevant scenarios (Section~\ref{sec:method-data-validation}). 
Using the synthesized scenario descriptions, {\tool} aligns an LLM with the captured traffic distributions
(\Cref{sec:method-align}).

\subsection{Data Synthesis}
\label{sec:method-extract}
To generate real-world aligned scenarios, {\tool} first captures the real-world traffic distributions in the target region.
We use driving videos as the data source because they are widely available online and cost-effective to collect.
From these videos, {\tool} automatically curates traffic scenario representations that reflect the underlying real-world traffic distributions.

\paragraph{Video Collection and Preprocessing.}
To demonstrate the generalizability of {\tool}, we collect {\numdataset} video datasets covering geographically diverse regions and use {\tool} to generate six synthetic datasets for alignment.
These six regions include two metropolitan areas (Los Angeles and New York City), two national parks (Yosemite and Yellowstone), a small town in Pennsylvania, United States, and a country-scale dataset from Switzerland encompassing diverse driving environments.
In total, we collect 261 first-person perspective scenic driving videos from YouTube~\cite{youtube}. 
{\tool} uniformly samples frames from each video at one frame every 15 seconds (fps=$1/15$) to capture diverse traffic scenarios while minimizing near-duplicate frames.
In-the-wild videos often contain irrelevant frames, such as titles, introductory screens, or auxiliary screens, which can result in invalid traffic scenario extraction. 
Scenarios synthesized based on these irrelevant frames will be systematically filtered by the semantic checkers in a later stage, as detailed in Section~\ref{sec:method-data-validation}.

\paragraph{Traffic Scenario Description Synthesis.}
For each selected frame, {\tool} employs a multimodal LLM to extract information about the depicted traffic scenario and synthesize a description in natural language.
We use GPT-4.1~nano (\texttt{gpt-4.1-nano-2025-04-14})~\cite{gpt-4.1-nano} as the default model in {\tool}.
Since LLM output quality is highly sensitive to prompt design, we carefully construct a compositional prompt that integrates several effective prompting strategies, including role-playing~\cite{shanahan2023roleplay}, step-by-step instructions, chain-of-thought reasoning~\cite{wei2022chain}, and in-context few-shot learning~\cite{brown2020language}.
The prompt includes four sections.
First, the LLM is assigned the role of a traffic domain expert. 
Second, the task is introduced with step-by-step instructions using a chain-of-thought strategy.
Third, we specify the output schema, which includes an intermediate reasoning trace and the natural language description of the traffic scenario.
Finally, the prompt includes an in-context example that familiarizes the LLM with both the input image and the expected output format.
Details on prompt design are available in {\supp}.

Specifically, the multimodal LLM is instructed to describe a traffic scenario with three parts. First, it needs to describe the road network and environment, including area type (urban or rural), road type (e.g., intersection, roundabout), directionality (one-way or two-way), number of lanes, weather, and time of day.
Then, it should provide contextual details, such as traffic density, roadside context, and the presence of emergency vehicles or road construction. Finally, it should also describe the actors, including the ego vehicle and the surrounding non-player character (NPC) vehicles and pedestrians. The LLM is instructed to describe the position and behavior of each actor. 
Examples of natural language descriptions are provided in {\supp}.

\subsection{Data Validation}
\label{sec:method-data-validation}
During the data synthesis stage, LLM hallucinations may produce semantically incomplete scenario representations.
In addition, in-the-wild input videos often contain irrelevant frames, such as titles, introductory screens, or auxiliary screens, which result in invalid scenario representations.
To address these potential issues and improve data quality for LLM alignment, {\tool} performs semantic validation on the synthesized traffic scenario descriptions.

\input{figs/fig_dual_representation}

\paragraph{A symbolic DSL as proxy.} 
Directly analyzing the semantics of the natural language descriptions is challenging and unreliable because such NL descriptions are unstructured and sometimes ambiguous. 
We therefore translate each description into a formal representation in a domain-specific language (DSL) for systematic analysis and rigorous checking.
We adopt an existing DSL design~\cite{DBLP:journals/tse/DengTYZZZ25}, since it provides a concise syntax grammar with sufficient coverage of various elements in a traffic scenario. 
Details of this DSL are provided in {\supp}.

{\tool} translates natural language descriptions into formal DSL representations using GPT-5 (\texttt{gpt-5-2025-08-07})~\cite{openai2025gpt5}.
The LLM prompt starts by configuring the LLM as a traffic domain expert and instructs it to identify the compositional elements of the traffic scenario step by step.
It then presents the definition of the DSL grammar.
The prompt ends with two in-context examples, each pairing a natural language description with its corresponding DSL representation.
The DSL's context-free grammar requires exact symbol usage.
{\tool} ensures that the translated DSL representation is well-formed by checking its syntax correctness based on the DSL grammar. For example, if it detects an out-of-vocabulary symbol, {\tool} will prompt the LLM again with the syntax error message and ask the LLM to refine it based on the grammar.
Figure~\ref{fig:dual-representation} illustrates a natural language description of a traffic scenario and its corresponding DSL representation.

\paragraph{Semantic validation.}
Using the formal DSL representation, {\tool} systematically validates each scenario representation for semantic completeness and validity.
For each DSL element, {\tool} verifies that the essential attributes are specified.
When multiple essential elements are missing (e.g., time of day and weather), this usually indicates that the source frame conveys little information related to traffic scenarios, so the \emph{semantic checker} marks it as a non-scenario frame and discards it.
If partial attributes of certain elements are missing (e.g., unspecified locations or behaviors for actors), the \emph{semantic checker} marks the scenario description as incomplete and initiates a self-refinement loop.

\paragraph{Self-refinement.}
{\tool} closes the loop between data synthesis and validation with a lightweight self-refinement step.
When a scenario is identified as valid but semantically incomplete, {\tool} appends the diagnostic messages to the original prompt and asks the \emph{scenario extractor} to regenerate the description.
Thus, {\tool} selectively refines recoverable scenarios and discards only semantically irrelevant ones, which helps capture the original traffic distribution of the target region.

\subsection{LLM Alignment}
\label{sec:method-align}
To align the LLM's knowledge with real-world traffic distributions, {\tool} fine-tunes the LLM using the traffic scenario descriptions synthesized from real-world driving videos.
We use Llama-3.2-3B-Instruct~\cite{llama3.2} as the default generator model in {\tool}.
{\tool} performs standard supervised fine-tuning (SFT) for the LLM with a next-token cross-entropy objective, computing loss only on assistant responses by masking user and system tokens.
{\tool} applies parameter-efficient fine-tuning (PEFT) via low-rank adaptation (LoRA) to attention and multilayer perceptron (MLP) projection layers.
Training runs for $60$ steps with a learning rate of $2e-4$.
The aligned LLM can be prompted to generate traffic scenario descriptions consistent with real-world traffic distribution.
Further details on LLM alignment, statistics of the {\numdataset} {\tool}-generated alignment datasets, and 
regional behavior differences
are provided in {\supp}.

%% file: figs/fig_pipeline.tex
\begin{figure*}[ht]
  \centering
\includegraphics[width=\linewidth]{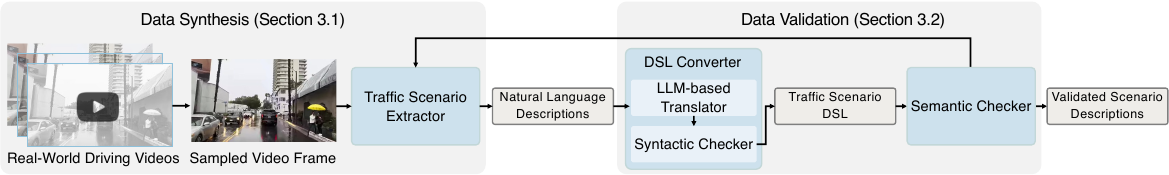}
\caption{Overview of {\tool}'s data synthesis pipeline.
}
\label{fig:pipeline}
\end{figure*}

%% file: figs/fig_dual_representation.tex
\begin{figure*}[ht]
  \centering
  \begin{subfigure}{0.33\textwidth}
    \includegraphics[width=\linewidth]{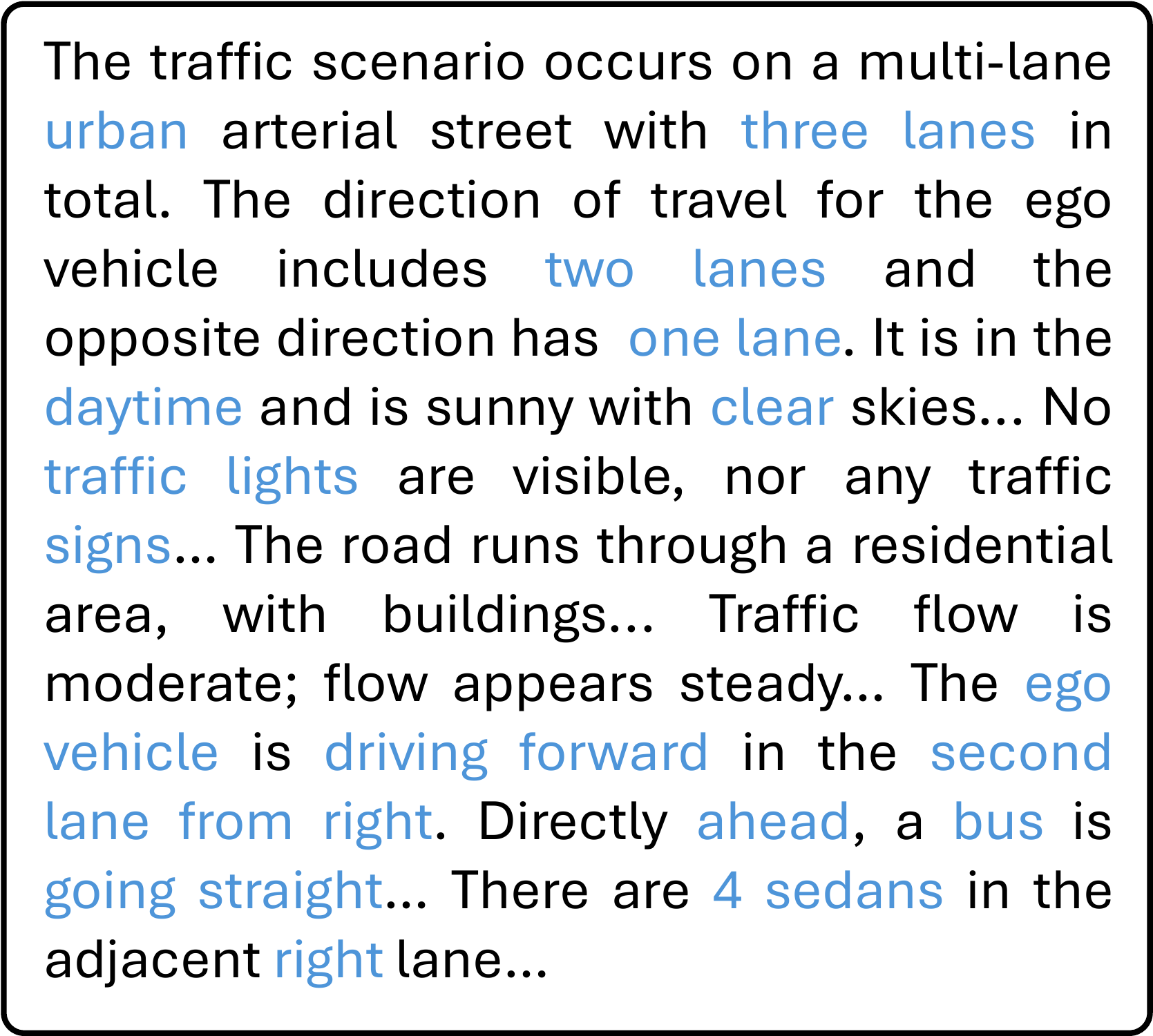}
    \caption{Natural language description.}
    \label{fig:dual-nl}
  \end{subfigure}
  \hfill
  \begin{subfigure}{0.66\textwidth}
    \includegraphics[width=\linewidth]{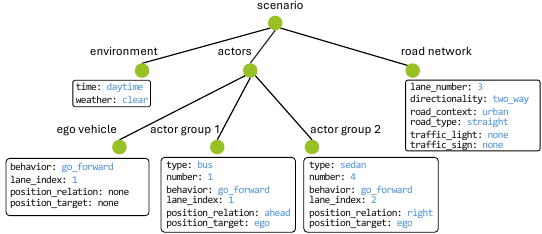}
    \caption{Symbolic representation using the domain-specific language~\cite{DBLP:journals/tse/DengTYZZZ25}.}
    \label{fig:dual-dsl}
  \end{subfigure}
\caption{Example natural language description of a traffic scenario and its translated domain-specific language representation for data validation. The natural language description is simplified for visualization.}
\label{fig:dual-representation}
\end{figure*}

%% file: sec/04_experiments.tex
\section{Experiments}
\label{sec:exp}

To comprehensively evaluate {\tool}, we conduct three experiments.
First, we evaluate the effectiveness of {\tool}-generated scenarios in testing {\numads} autonomous driving models (Section~\ref{sec:exp_adstest}).
Second, we fine-tune these driving models on {\tool}-generated scenarios and test them on a held-out set to evaluate {\tool}'s effectiveness in improving driving model performance (Section~\ref{sec:exp_ft}).
Third, we conduct a qualitative study
to analyze how well {\tool}-generated scenarios align with real-world traffic distributions (Section~\ref{sec:exp_vis}).

\input{tables/tab_fuzz_ads_test.tex}

\subsection{Experimental Setup}
{\tool}-generated traffic scenarios are represented as natural language descriptions, which are difficult to evaluate directly. 
To quantitatively assess their real-world usefulness, we convert these descriptions into Scenic~\cite{DBLP:journals/ml/FremontKDGYSS23} scripts.
These Scenic scripts automatically configure the road network, weather, time, and the positions and behaviors of surrounding vehicles and pedestrians in the CARLA simulator~\cite{pmlr-v78-dosovitskiy17a}.
In simulation experiments, the ego vehicle is controlled by a reinforcement learning-based autonomous driving model. 
All quantitative evaluations are performed using the SafeBench platform~\cite{10.5555/3600270.3602131}.

\paragraph{Simulation script conversion.}
To convert {\tool}-generated traffic scenario descriptions into Scenic~\cite{DBLP:journals/ml/FremontKDGYSS23} scripts, we follow the algorithm introduced in TrafficComposer~\cite{trafficcomposer}.
TrafficComposer uses an LLM to extract traffic information from its inputs, post-processes missing fields, and applies a rule-based algorithm to compose simulation scripts. 
We build on its implementation with two modifications. 
(i) TrafficComposer accepts multimodal inputs, whereas we use text-only (single-modal) natural language descriptions; therefore, we discard TrafficComposer's visual input processing pipeline.
(ii) TrafficComposer proposes a compositional simulation script converter but does not implement it for Scenic~\cite{DBLP:journals/ml/FremontKDGYSS23}, so we follow its algorithm design and implement a rule-based Scenic script converter.

\paragraph{Scenario generation baselines.}
For evaluation, we compare {\tool} with {\numbaselines} representative scenario generation baselines and {\numllmbaseline} LLM baselines.

The strongest baseline is the state-of-the-art method, \textbf{ChatScene}~\cite{Zhang_2024_CVPR}, which prompts an LLM to generate textual descriptions of traffic scenarios and uses a predefined retrieval database to translate them into executable Scenic scripts.
We also include two adversary-based methods, \textbf{Learning-to-Collide}~\cite{ding2020learning} and \textbf{AdvSim}~\cite{wang2021advsim}, which challenge driving models by perturbing the trajectories or initial configurations of surrounding vehicles and pedestrians. 
In addition, we include two rule-based methods, \textbf{Carla Scenario Generator}~\cite{carla2019scenariorunner} and \textbf{Adversarial Trajectory Optimization}~\cite{zhang2022adversarial}, which construct scenarios consistent with predefined real-world traffic rules and physical principles.

To ablate the effect of aligning LLM knowledge with real-world traffic distributions, we construct {\tool} variants that do not perform LLM alignment.
Instead, the LLM is prompted only to generate scenario descriptions for the specified location.
We experiment with {\numllmbaseline} state-of-the-art LLMs, GPT-5 (\texttt{gpt-5-2025-08-07})~\cite{openai2025gpt5}, GPT-4o (\texttt{gpt-4o-2024-11-20})~\cite{openai2024gpt4o}, Claude Sonnet 4 (\texttt{claude-sonnet-4@20250514})~\cite{anthropic2025claude_sonnet4}, DeepSeek-V3 (\texttt{DeepSeek-V3-0324})~\cite{deepseekv3}, and Qwen3 (\texttt{Qwen3-32B-FP8})~\cite{qwen3_32b_fp8_2025, yang2025qwen3}.
Prompt details are provided in the {\supp}.

\paragraph{Autonomous driving models.}
We conduct experiments using three widely adopted deep reinforcement learning (RL) algorithms to control the ego vehicle: Proximal Policy Optimization (PPO)~\cite{schulman2017proximal}, an on-policy stochastic algorithm; Soft Actor-Critic (SAC)~\cite{haarnoja2018soft}, an off-policy stochastic algorithm; and Twin Delayed Deep Deterministic Policy Gradient (TD3)~\cite{fujimoto2018addressing}, a deterministic off-policy algorithm.

In our experiments, the observation space of the RL agents consists of four features: distance to the next waypoint, longitudinal speed, angular speed, and a detection signal for front-facing vehicles.

\paragraph{Quantitative Metrics.}
We measure the performance of driving models using $10$ metrics provided by the SafeBench platform~\cite{10.5555/3600270.3602131} across three performance levels: the Safety level (collision rate and adherence to traffic signals), the Functionality level (route adherence and completion), and the Etiquette level (smoothness of driving and lane discipline). 
An overall score is computed by aggregating all $10$ metrics to provide a holistic assessment of performance.
Our analysis primarily focuses on the collision rate and the overall score.

\input{tables/tab_ablation}

\subsection{Effectiveness in Testing Driving Models}
\label{sec:exp_adstest}
We evaluate the effectiveness of {\tool} in generating challenging and safety-critical traffic scenarios for autonomous driving models.
We compare {\tool} with the state-of-the-art scenario generation method ChatScene~\cite{Zhang_2024_CVPR}, two adversary-based methods~\cite{ding2020learning, wang2021advsim}, two rule-based methods~\cite{carla2019scenariorunner, zhang2022adversarial}, and {\numllmbaseline} LLM baselines.

\paragraph{Experiment setup.}
The evaluation follows the experimental settings of the state-of-the-art baseline ChatScene~\cite{Zhang_2024_CVPR}.
The evaluation pipeline first mutates each generated traffic scenario to identify challenging variants.
The procedure measures scenario difficulty with a surrogate ego vehicle controlled by SAC, based on collision occurrence and the \emph{overall score} reported from the SafeBench platform~\cite{10.5555/3600270.3602131}.
For each traffic scenario, the procedure performs 50 rounds of search and applies parameter updates every 10 steps. 
From the 50 variants, the two most challenging are retrained for evaluation.
The scenarios are evaluated using ego vehicles trained with {\numads} driving models---PPO, SAC, and TD3. 
ChatScene~\cite{Zhang_2024_CVPR} starts with 40 initial traffic scenarios.
To ensure a fair comparison, {\tool} and the {\numllmbaseline} unaligned LLM baselines are evaluated under identical settings, randomly generating 40 scenarios as starting points to match ChatScene~\cite{Zhang_2024_CVPR}.
For a comprehensive evaluation, we evaluate {\numdataset} instances of {\tool}, each constructed for a distinct geographic region.
This quantifies the effectiveness of scenario generation methods in testing autonomous driving models.

For the two adversary-based methods, Learning-to-Collide~\cite{ding2020learning} and AdvSim~\cite{wang2021advsim}, and the two rule-based methods, Carla Scenario Generator~\cite{carla2019scenariorunner} and Adversarial Trajectory Optimization~\cite{zhang2022adversarial}, we adopt ChatScene's implementation.

\input{tables/tab_tune_ads}

\paragraph{Evaluation results.}
Table~\ref{tab:fuzz} presents the average test results of {\numads} driving models in scenarios generated by {\tool} and {\numbaselines} representative baselines. 
All {\tool} instances across the {\numdataset} geographically diverse regions consistently outperform existing baseline methods on almost all metrics.
Specifically, {\tool} generates significantly more prone-to-crash traffic scenarios, achieving a \result{2.7\%}--{\crimprove} increase in \emph{collision rate} compared with the best-performing baseline.
In terms of driving models' overall performance, {\tool} induces a relative reduction in the average \emph{overall score}, ranging from \result{5.0\%} to {\osimprove} compared with the strongest baseline.
These results indicate that {\tool} generates more prone-to-crash and challenging traffic scenarios, thereby enabling more effective testing of autonomous driving models.

An \textbf{ablation study} evaluates the effect of aligning LLM's knowledge with real-world traffic distribution.
\Cref{tab:ablation} presents the average test results of {\numads} driving models in scenarios generated by {\tool} and {\numllmbaseline} LLM baselines.
Due to space limitations, we only report results from the \emph{Los Angeles} region.
{\tool} consistently outperforms the LLM baselines across all $11$ metrics, 
with a \result{3.9\%}--\result{12.9\%} increase in \emph{collision rate} and a \result{6.9\%}--\result{10.0\%} reduction in \emph{overall score}.
These results indicate that alignment with real-world distributions enables {\tool} to generate more challenging traffic scenarios, thereby enabling more effective autonomous driving model testing.

In conclusion, {\tool}-generated scenarios are effective in testing autonomous driving models. 
The ablation study shows that aligning LLM knowledge with the real-world traffic distribution generates more challenging scenarios.
The results highlight the potential of {\tool} to establish new benchmarks for testing the safety and robustness of autonomous driving models.

\subsection{Effectiveness in Improving Driving Models}
\label{sec:exp_ft}
This experiment evaluates the effectiveness of {\tool} in improving autonomous driving model performance.
We fine-tune the {\numads} driving models on scenarios generated by different methods and test them on a held-out composite test set to quantify the performance improvement.

\paragraph{Experiment setup.}
For consistency, we fine-tune the same {\numads} autonomous driving models---PPO, SAC, and TD3.
For each scenario generation method, we use $48$ scenarios for model fine-tuning (from $40 \times 2$ scenario variants with a $3/5$ split) and evaluate on the remaining $32$ ($2/5$) scenarios, aggregated across all generation methods to form a held-out composite test set.
We follow the experiment settings of ChatScene~\cite{Zhang_2024_CVPR} for model fine-tuning and evaluation.
For each driving model, we initialize from the SafeBench~\cite{10.5555/3600270.3602131} provided model checkpoint and fine-tune it for $500$ epochs with a learning rate of $0.0001$.
We evaluate performance every $50$ epochs and report the best result achieved across these evaluations.
Given the high cost of agent fine-tuning, we do not include the four baselines~\cite{ding2020learning, wang2021advsim, carla2019scenariorunner, zhang2022adversarial} that are significantly weaker in model testing (\Cref{sec:exp_adstest}) and focus on comparison with the state-of-the-art baseline, ChatScene~\cite{Zhang_2024_CVPR}.

\paragraph{Evaluation Results.}
Table~\ref{tab:tune} presents the average test results of {\numads} autonomous driving models after fine-tuning on scenarios generated by ChatScene~\cite{Zhang_2024_CVPR} and {\tool}.
Models fine-tuned on {\tool}-generated scenarios consistently achieve the best performance on almost all $11$ metrics across the {\numdataset} geographically diverse regions.
Specifically, we observe a significant \result{21.0\%}--{\ftcrimprovefromraw} reduction in collision rate compared with the original driving model without fine-tuning, and a relative improvement in overall score of \result{22.4\%}--\result{49.7\%}.
Compared with the state-of-the-art baseline, ChatScene~\cite{Zhang_2024_CVPR}, {\tool} achieves a further reduction in collision rate by \result{2.0\%}--\result{3.7\%} and a further improvement in overall score by \result{5.1\%}--\result{27.9\%}.

In conclusion, {\tool}-generated scenarios are highly effective in improving autonomous driving model performance.
Fine-tuning driving models on the real-world aligned scenarios generated by {\tool} significantly improves robustness and overall performance.

\input{figs/fig_vis_baselines}

\subsection{Qualitative Study on Distribution Alignment}
\label{sec:exp_vis}
This qualitative study assesses whether the generated traffic scenarios align with the real-world traffic distribution.

\paragraph{Experiment Setup.}
We generate scenarios for a target region using {\tool} and {\numllmbaseline} unaligned LLM baselines and compare them with real-world scenarios extracted from driving videos.
The real-world samples are excluded from the LLM alignment for {\tool}.
We embed the scenario descriptions using the Sentence Transformers \texttt{all-mpnet-base-v2} model~\cite{reimers-2019-sentence-bert, all-mpnet-base-v2}, then project the embeddings into two-dimensional space using Uniform Manifold Approximation and Projection (UMAP)~\cite{mcinnes2018umap} for visualization. 
For the best visualization, we generate $40$ scenarios per method.
Due to page limit, we only include the results of three regions---Los Angeles, Yellowstone, and the small town area in Pennsylvania, USA.
Full results are reported in {\supp}.

\paragraph{Evaluation Results.}
Figure~\ref{fig:vis-baselines} visualizes scenarios generated by {\tool} and the {\numllmbaseline} unaligned LLM baselines alongside real-world cases in the embedding space.
Across the three geographically diverse regions,
{\tool}-generated scenarios consistently co-locate with and overlap the \emph{Real World} cluster.
In contrast, GPT-4o, Claude-Sonnet-4, DeepSeek-V3, and Qwen3 form clusters displaced from the real-world region with barely any overlap, and GPT-5 forms an isolated cluster far from the real-world cases.
In conclusion, the observed co-location and overlap indicate that {\tool} captures the dominant modes of the real-world traffic distribution and reduces distribution shift relative to unaligned LLM baselines.

%% file: tables/tab_fuzz_ads_test.tex
\begin{table*}[t]
\centering
\caption{
\textbf{Effectiveness of scenarios in autonomous driving model testing.}
This table presents the average test results on {\numads} distinct autonomous driving models using scenarios generated by five baselines and by {\tool} on six geographically diverse regions. 
CR: collision rate, RR: frequency of running red lights, SS: frequency of running stop signs, OR: average distance driven out of road, RF: route following stability, Comp: average percentage of route completion, TS: average time spent to complete the route, ACC: average acceleration, YV: average yaw velocity, LI: frequency of lane invasion, OS: overall score.
$\uparrow$/$\downarrow$: higher/lower values indicate more effective in testing the driving models.
\textbf{Values in bold} indicate the best performance; \colorbox{myresultgreen}{values shaded green} indicate that {\tool} outperforms all baselines.
}
\label{tab:fuzz}
\resizebox{0.92\linewidth}{!}{%

\begin{tabular}{l|cccc|ccc|ccc|c}
\toprule
\multicolumn{1}{c|}{} & \multicolumn{4}{c|}{\textbf{Safety Level}} & \multicolumn{3}{c|}{\textbf{Functionality Level}} & \multicolumn{3}{c|}{\textbf{Etiquette Level}} &  \\
\multicolumn{1}{c|}{\multirow{-2}{*}{\textbf{Method}}} & CR $\uparrow$ & RR $\uparrow$ & SS $\uparrow$ & OR $\uparrow$ & RF $\downarrow$ & Comp $\downarrow$ & TS $\uparrow$ & ACC $\uparrow$ & YV $\uparrow$ & LI $\uparrow$ & \multirow{-2}{*}{OS $\downarrow$} \\ \midrule
Learning-to-collide~\cite{ding2020learning} & 0.584 & 0.326 & 0.158 & 0.032 & 0.894 & 0.731 & 0.216 & 0.211 & 0.243 & 0.112 & 0.619 \\
AdvSim~\cite{wang2021advsim} & 0.586 & 0.300 & 0.160 & 0.025 & 0.891 & 0.745 & 0.261 & 0.203 & 0.245 & 0.127 & 0.620 \\
Carla Scenario Generator~\cite{carla2019scenariorunner} & 0.676 & 0.313 & 0.161 & 0.036 & 0.890 & 0.741 & 0.244 & 0.215 & 0.243 & 0.131 & 0.573 \\
Adv. Trajectory Optim.~\cite{zhang2022adversarial} & 0.627 & 0.312 & 0.158 & 0.028 & 0.893 & 0.726 & 0.279 & 0.219 & 0.248 & 0.137 & 0.596 \\
ChatScene~\cite{Zhang_2024_CVPR} & 0.825 & 0.191 & 0.142 & 0.035 & 0.861 & 0.537 & 0.213 & 0.706 & 0.533 & 0.228 & 0.481 \\ \midrule
{\tool} (Los Angeles) & \cellcolor[HTML]{D5E5CF}\textbf{0.933} & 0.316 & \cellcolor[HTML]{D5E5CF}\textbf{0.283} & \cellcolor[HTML]{D5E5CF}0.091 & \cellcolor[HTML]{D5E5CF}0.713 & \cellcolor[HTML]{D5E5CF}0.524 & \cellcolor[HTML]{D5E5CF}\textbf{0.491} & \cellcolor[HTML]{D5E5CF}0.766 & \cellcolor[HTML]{D5E5CF}0.601 & \cellcolor[HTML]{D5E5CF}0.319 & \cellcolor[HTML]{D5E5CF}0.405 \\
{\tool} (New York City) & \cellcolor[HTML]{D5E5CF}0.923 & \cellcolor[HTML]{D5E5CF}\textbf{0.466} & \cellcolor[HTML]{D5E5CF}0.281 & \cellcolor[HTML]{D5E5CF}0.037 & \cellcolor[HTML]{D5E5CF}\textbf{0.700} & \cellcolor[HTML]{D5E5CF}0.527 & \cellcolor[HTML]{D5E5CF}0.391 & \cellcolor[HTML]{D5E5CF}0.793 & \cellcolor[HTML]{D5E5CF}0.541 & \cellcolor[HTML]{D5E5CF}0.357 & \cellcolor[HTML]{D5E5CF}0.319 \\
{\tool} (Yellowstone) & \cellcolor[HTML]{D5E5CF}0.909 & \cellcolor[HTML]{D5E5CF}0.456 & \cellcolor[HTML]{D5E5CF}0.221 & \cellcolor[HTML]{D5E5CF}\textbf{0.104} & \cellcolor[HTML]{D5E5CF}0.708 & \cellcolor[HTML]{D5E5CF}\textbf{0.438} & \cellcolor[HTML]{D5E5CF}0.298 & \cellcolor[HTML]{D5E5CF}\textbf{0.809} & 0.476 & \cellcolor[HTML]{D5E5CF}0.355 & \cellcolor[HTML]{D5E5CF}\textbf{0.310} \\
{\tool} (Yosemite) & \cellcolor[HTML]{D5E5CF}0.903 & \cellcolor[HTML]{D5E5CF}0.420 & \cellcolor[HTML]{D5E5CF}0.179 & \cellcolor[HTML]{D5E5CF}0.039 & \cellcolor[HTML]{D5E5CF}0.740 & 0.582 & \cellcolor[HTML]{D5E5CF}0.477 & \cellcolor[HTML]{D5E5CF}0.756 & \cellcolor[HTML]{D5E5CF}\textbf{0.660} & 0.218 & \cellcolor[HTML]{D5E5CF}0.457 \\
{\tool} (Small Towns) & \cellcolor[HTML]{D5E5CF}0.852 & 0.186 & 0.149 & \cellcolor[HTML]{D5E5CF}{\color[HTML]{000000} 0.038} & \cellcolor[HTML]{D5E5CF}0.796 & 0.561 & 0.266 & \cellcolor[HTML]{D5E5CF}0.735 & 0.517 & \cellcolor[HTML]{D5E5CF}0.283 & \cellcolor[HTML]{D5E5CF}0.437 \\
{\tool} (Switzerland) & \cellcolor[HTML]{D5E5CF}0.919 & \cellcolor[HTML]{D5E5CF}0.379 & \cellcolor[HTML]{D5E5CF}0.201 & \cellcolor[HTML]{D5E5CF}{\color[HTML]{000000} 0.041} & \cellcolor[HTML]{D5E5CF}0.723 & \cellcolor[HTML]{D5E5CF}0.514 & \cellcolor[HTML]{D5E5CF}0.383 & \cellcolor[HTML]{D5E5CF}0.761 & 0.514 & \cellcolor[HTML]{D5E5CF}\textbf{0.393} & \cellcolor[HTML]{D5E5CF}0.429 \\ \bottomrule
\end{tabular}

}

\end{table*}

%% file: tables/tab_ablation.tex
\begin{table*}[thb]
\centering
\caption{
\textbf{Ablation study on LLM alignment: scenario effectiveness in autonomous driving model testing.}
This table presents the average test results on {\numads} distinct autonomous driving models using traffic scenarios in \textbf{Los Angeles}, generated by our {\tool} and {\numllmbaseline} LLMs without knowledge alignment. 
CR: collision rate, RR: frequency of running red lights, SS: frequency of running stop signs, OR: average distance driven out of road, RF: route following stability, Comp: average percentage of route completion, TS: average time spent to complete the route, ACC: average acceleration, YV: average yaw velocity, LI: frequency of lane invasion, OS: overall score, 
$\uparrow$/$\downarrow$: higher/lower values indicate more effective in testing the driving models.
\textbf{Values in bold} indicate the best performance.
}
\label{tab:ablation}
\resizebox{0.8\linewidth}{!}{%

\begin{tabular}{l|cccc|ccc|ccc|c}
\toprule
\multicolumn{1}{c|}{\multirow{2}{*}{\textbf{Method}}} & \multicolumn{4}{c|}{\textbf{Safety Level}} & \multicolumn{3}{c|}{\textbf{Functionality Level}} & \multicolumn{3}{c|}{\textbf{Etiquette Level}} & \multirow{2}{*}{OS $\downarrow$} \\
\multicolumn{1}{c|}{} & CR $\uparrow$ & RR $\uparrow$ & SS $\uparrow$ & OR $\uparrow$ & RF $\downarrow$ & Comp $\downarrow$ & TS $\uparrow$ & ACC $\uparrow$ & YV $\uparrow$ & LI $\uparrow$ &  \\ \midrule
GPT-5 & 0.889 & 0.291 & 0.163 & 0.034 & 0.784 & 0.557 & 0.251 & 0.728 & 0.529 & 0.245 & 0.435 \\
GPT-4o & 0.874 & 0.284 & 0.165 & 0.035 & 0.791 & 0.549 & 0.268 & 0.729 & 0.531 & 0.242 & 0.441 \\
Claude Sonnet 4 & 0.894 & 0.243 & 0.187 & 0.037 & 0.786 & 0.539 & 0.259 & 0.714 & 0.541 & 0.241 & 0.438 \\
DeepSeek-V3 & 0.814 & 0.273 & 0.159 & 0.033 & 0.806 & 0.583 & 0.238 & 0.724 & 0.527 & 0.239 & 0.447 \\
Qwen3 & 0.804 & 0.278 & 0.153 & 0.035 & 0.796 & 0.571 & 0.207 & 0.719 & 0.522 & 0.219 & 0.450 \\

\textbf{\tool} & \textbf{0.933} & \textbf{0.316} & \textbf{0.283} & \textbf{0.091} & \textbf{0.713} & \textbf{0.524} & \textbf{0.491} & \textbf{0.766} & \textbf{0.601} & \textbf{0.319} & \textbf{0.405} \\ \bottomrule
\end{tabular}

}

\vspace{-1em}
\end{table*}

%% file: tables/tab_tune_ads.tex
\begin{table*}[t]
\centering
\caption{
\textbf{Effectiveness of scenarios in improving autonomous driving model performance}: 
This table presents the average test results for {\numads} autonomous driving models after fine-tuning. The driving models are fine-tuned on a subset of scenarios generated by each method and tested on a composite held-out set aggregating test scenarios from all methods. 
CR: collision rate, RR: frequency of running red lights, SS: frequency of running stop signs, OR: average distance driven out of road, RF: route following stability, Comp: average percentage of route completion, TS: average time spent to complete the route, ACC: average acceleration, YV: average yaw velocity, LI: frequency of lane invasion, OS: overall score.
$\uparrow$/$\downarrow$: higher/lower the more effective in improving driving model performance.
\textbf{Values in bold} indicate the best performance; \colorbox{myresultgreen}{values shaded green} indicate that {\tool} outperforms all baselines.
}
\label{tab:tune}

\resizebox{0.94\linewidth}{!}{%

\begin{tabular}{ll|cccc|ccc|ccc|c}
\toprule
\multicolumn{1}{c}{} & \multicolumn{1}{c|}{} & \multicolumn{4}{c|}{\textbf{Safety Level}} & \multicolumn{3}{c|}{\textbf{Functionality Level}} & \multicolumn{3}{c|}{\textbf{Etiquette Level}} &  \\
\multicolumn{1}{c}{\multirow{-2}{*}{\textbf{Location}}} & \multicolumn{1}{c|}{\multirow{-2}{*}{\textbf{Method}}} & CR $\downarrow$ & RR $\downarrow$ & SS $\downarrow$ & OR $\downarrow$ & RF $\uparrow$ & Comp $\uparrow$ & TS $\downarrow$ & ACC $\downarrow$ & YV $\downarrow$ & LI $\downarrow$ & \multirow{-2}{*}{OS $\uparrow$} \\ \midrule
 & w/o fine-tuning & 0.968 & 0.603 & 0.149 & 0.116 & 0.093 & 0.291 & 0.741 & 0.771 & 0.861 & 0.701 & 0.296 \\
 & ChatScene~\cite{Zhang_2024_CVPR} & 0.733 & 0.461 & 0.122 & 0.009 & 0.631 & 0.304 & 0.628 & 0.709 & 0.533 & 0.581 & 0.301 \\
\multirow{-3}{*}{Los Angeles} & \textbf{\tool} & \cellcolor[HTML]{D5E5CF}\textbf{0.701} & \cellcolor[HTML]{D5E5CF}\textbf{0.388} & \cellcolor[HTML]{D5E5CF}\textbf{0.000} & \cellcolor[HTML]{D5E5CF}\textbf{0.000} & \cellcolor[HTML]{D5E5CF}\textbf{0.793} & \cellcolor[HTML]{D5E5CF}\textbf{0.598} & \cellcolor[HTML]{D5E5CF}\textbf{0.256} & \cellcolor[HTML]{D5E5CF}\textbf{0.613} & \cellcolor[HTML]{D5E5CF}\textbf{0.516} & \cellcolor[HTML]{D5E5CF}\textbf{0.498} & \cellcolor[HTML]{D5E5CF}\textbf{0.385} \\ \midrule
 & w/o fine-tuning & 0.973 & 0.591 & 0.168 & 0.108 & 0.114 & 0.296 & 0.728 & 0.734 & 0.858 & 0.696 & 0.299 \\
 & ChatScene~\cite{Zhang_2024_CVPR} & 0.734 & 0.384 & \textbf{0.000} & \textbf{0.000} & 0.361 & 0.442 & 0.643 & 0.592 & 0.801 & 0.509 & 0.314 \\
\multirow{-3}{*}{New York City} & \textbf{\tool} & \cellcolor[HTML]{D5E5CF}\textbf{0.710} & \cellcolor[HTML]{D5E5CF}\textbf{0.378} & \cellcolor[HTML]{D5E5CF}\textbf{0.000} & \cellcolor[HTML]{D5E5CF}\textbf{0.000} & \cellcolor[HTML]{D5E5CF}\textbf{0.800} & \cellcolor[HTML]{D5E5CF}\textbf{0.595} & \cellcolor[HTML]{D5E5CF}\textbf{0.258} & \cellcolor[HTML]{D5E5CF}\textbf{0.568} & \cellcolor[HTML]{D5E5CF}\textbf{0.501} & \cellcolor[HTML]{D5E5CF}\textbf{0.399} & \cellcolor[HTML]{D5E5CF}\textbf{0.366} \\ \midrule
 & w/o fine-tuning & 0.953 & 0.633 & 0.231 & 0.102 & 0.108 & 0.301 & 0.733 & 0.741 & 0.842 & 0.652 & 0.310 \\
 & ChatScene~\cite{Zhang_2024_CVPR} & 0.768 & \textbf{0.410} & 0.060 & \textbf{0.000} & 0.753 & 0.528 & 0.413 & 0.671 & 0.591 & \textbf{0.377} & 0.405 \\
\multirow{-3}{*}{Yellowstone} & \textbf{\tool} & \cellcolor[HTML]{D5E5CF}\textbf{0.743} & 0.419 & \cellcolor[HTML]{D5E5CF}\textbf{0.000} & \cellcolor[HTML]{D5E5CF}\textbf{0.000} & \cellcolor[HTML]{D5E5CF}\textbf{0.767} & \cellcolor[HTML]{D5E5CF}\textbf{0.569} & \cellcolor[HTML]{D5E5CF}\textbf{0.385} & \cellcolor[HTML]{D5E5CF}\textbf{0.660} & \cellcolor[HTML]{D5E5CF}\textbf{0.570} & 0.410 & \cellcolor[HTML]{D5E5CF}\textbf{0.439} \\ \midrule
 & w/o fine-tuning & 0.965 & 0.609 & 0.204 & 0.133 & 0.103 & 0.305 & 0.731 & 0.730 & 0.854 & 0.663 & 0.304 \\
 & ChatScene~\cite{Zhang_2024_CVPR} & 0.641 & 0.459 & \textbf{0.000} & \textbf{0.002} & 0.104 & 0.498 & 0.366 & 0.724 & 0.593 & 0.461 & 0.427 \\
\multirow{-3}{*}{Yosemite} & \textbf{\tool} & \cellcolor[HTML]{D5E5CF}\textbf{0.604} & \cellcolor[HTML]{D5E5CF}\textbf{0.435} & \cellcolor[HTML]{D5E5CF}\textbf{0.000} & \cellcolor[HTML]{D5E5CF}\textbf{0.002} & \cellcolor[HTML]{D5E5CF}\textbf{0.841} & \cellcolor[HTML]{D5E5CF}\textbf{0.527} & \cellcolor[HTML]{D5E5CF}\textbf{0.352} & \cellcolor[HTML]{D5E5CF}\textbf{0.666} & \cellcolor[HTML]{D5E5CF}\textbf{0.538} & \cellcolor[HTML]{D5E5CF}\textbf{0.428} & \cellcolor[HTML]{D5E5CF}\textbf{0.455} \\ \midrule
 & w/o fine-tuning & 0.956 & 0.582 & 0.201 & 0.109 & 0.142 & 0.331 & 0.671 & 0.734 & 0.833 & 0.602 & 0.311 \\
 & ChatScene~\cite{Zhang_2024_CVPR} & 0.649 & 0.536 & \textbf{0.000} & 0.031 & 0.653 & \textbf{0.541} & 0.591 & 0.713 & 0.766 & \textbf{0.409} & 0.401 \\
\multirow{-3}{*}{Small Towns} & \textbf{\tool} & \cellcolor[HTML]{D5E5CF}\textbf{0.629} & \cellcolor[HTML]{D5E5CF}\textbf{0.296} & \cellcolor[HTML]{D5E5CF}\textbf{0.000} & \cellcolor[HTML]{D5E5CF}\textbf{0.029} & \cellcolor[HTML]{D5E5CF}\textbf{0.733} & 0.532 & \cellcolor[HTML]{D5E5CF}\textbf{0.348} & \cellcolor[HTML]{D5E5CF}\textbf{0.501} & \cellcolor[HTML]{D5E5CF}\textbf{0.588} & 0.426 & \cellcolor[HTML]{D5E5CF}\textbf{0.433} \\ \midrule
 & w/o fine-tuning & 0.954 & 0.596 & 0.109 & 0.098 & 0.116 & 0.295 & 0.704 & 0.728 & 0.846 & 0.688 & 0.303 \\
 & ChatScene~\cite{Zhang_2024_CVPR} & 0.653 & 0.431 & \textbf{0.000} & \textbf{0.000} & 0.682 & 0.392 & 0.461 & 0.639 & \textbf{0.532} & \textbf{0.409} & 0.389 \\
\multirow{-3}{*}{Switzerland} & \textbf{\tool} & \cellcolor[HTML]{D5E5CF}\textbf{0.625} & \cellcolor[HTML]{D5E5CF}\textbf{0.428} & \cellcolor[HTML]{D5E5CF}\textbf{0.000} & \cellcolor[HTML]{D5E5CF}\textbf{0.000} & \cellcolor[HTML]{D5E5CF}\textbf{0.699} & \cellcolor[HTML]{D5E5CF}\textbf{0.566} & \cellcolor[HTML]{D5E5CF}\textbf{0.391} & \cellcolor[HTML]{D5E5CF}\textbf{0.591} & 0.533 & 0.412 & \cellcolor[HTML]{D5E5CF}\textbf{0.409} \\ \bottomrule
\end{tabular}

}

\end{table*}

%% file: figs/fig_vis_baselines.tex
\begin{figure*}[ht]
  \centering
  \begin{subfigure}{0.32\textwidth}
    \includegraphics[width=\linewidth]{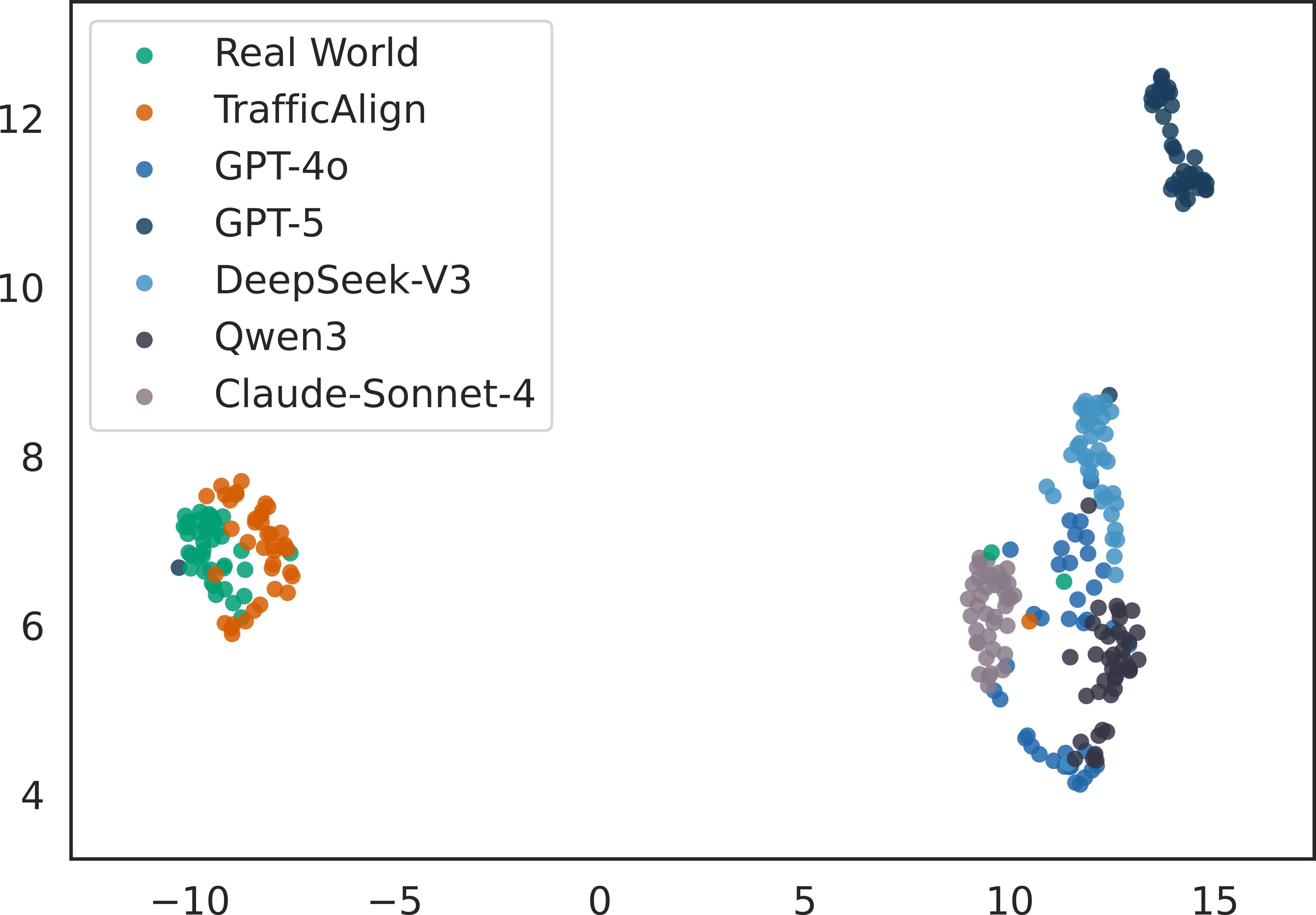}
    \caption{Los Angeles.}
    \label{fig:vis-baselines-la}
  \end{subfigure}
  \hfill
  \begin{subfigure}{0.32\textwidth}
    \includegraphics[width=\linewidth]{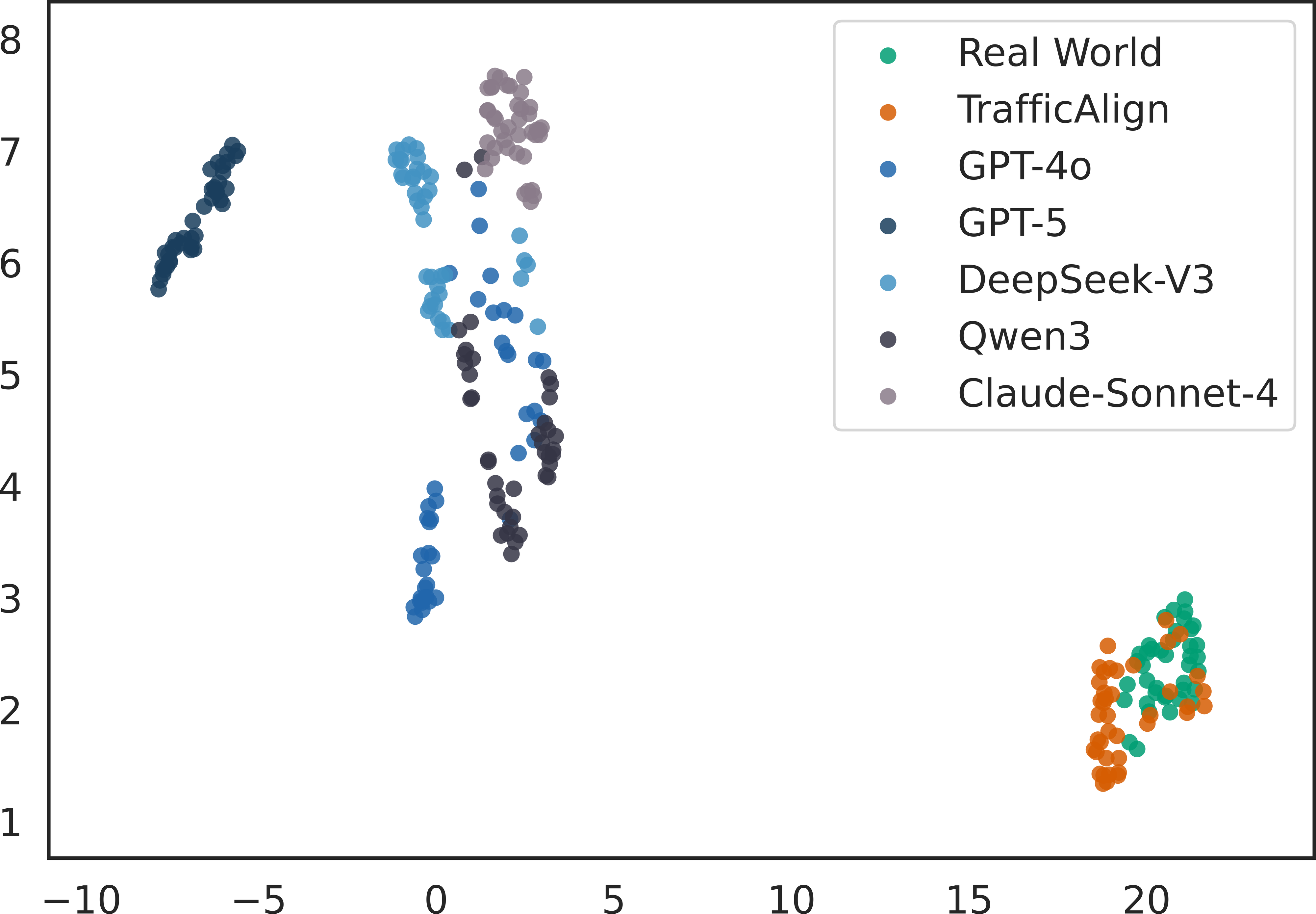}
    \caption{Yellowstone.}
    \label{fig:vis-baselines-yellowstone}
  \end{subfigure}
  \hfill
  \begin{subfigure}{0.32\textwidth}
    \includegraphics[width=\linewidth]{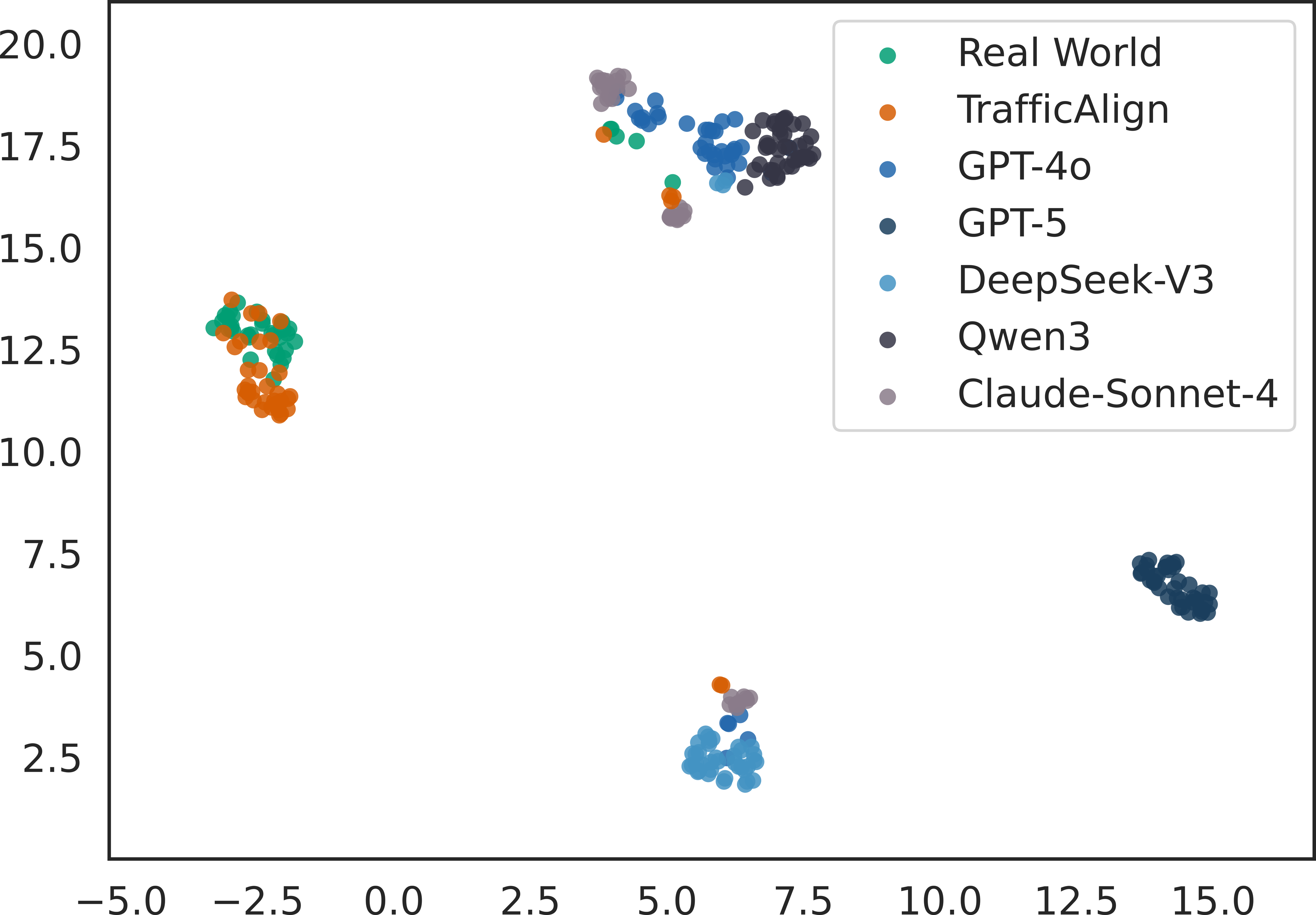}
    \caption{Small Towns in Pennsylvania, USA.}
    \label{fig:vis:vis-baselines-smalltowns}
  \end{subfigure}
  \caption{UMAP visualization of embeddings of traffic scenarios generated by {\tool}, {\numllmbaseline} unaligned LLM baselines, and real-world traffic scenarios.}
  \label{fig:vis-baselines}
\end{figure*}

%% file: sec/05_limitations.tex
\section{Limitations and Future Work}
\label{sec:limitations}
{\tool} automatically synthesizes and validates traffic scenarios for LLM alignment from in-the-wild driving videos, but has 
limitations that motivate future work. 
First, {\tool} uniformly samples frames from videos (Section~\ref{sec:method-extract}), keeping the scenario set 
representative of the target region, but can potentially miss rare traffic events.
Future versions can adopt distribution-aware or event-triggered sampling to better cover long-tail situations.
Second, the scenario extractor operates on a single frame per scenario and depends on a multimodal LLM to perceive visual information and infer traffic dynamics. 
This design keeps the pipeline simple, but can lose temporal cues for understanding actor behavior.
A promising future direction is to use video clips and specialized computer vision models to extract visual information while using LLMs for high-level reasoning.
However, we emphasize that {\tool} aims to propose a paradigm to automatically align LLMs with real-world traffic scenarios. 
Although these limitations may affect performance, they do not undermine the validity of the findings. 

%% file: sec/06_conclusion.tex
\section{Conclusion}
\label{sec:conclusion}
In this work, we introduce {\tool}, an automated framework that synthesizes traffic scenarios based on real-world driving videos, performs data validation, and aligns LLMs with the synthesized scenarios.
Our evaluation shows that traffic scenarios generated by {\tool} are effective in autonomous driving model testing, revealing up to {\crimprove} more collisions on average across three driving models compared with state-of-the-art methods.
Furthermore, fine-tuning these driving models with {\tool}-generated scenarios significantly reduced collision rates by {\ftcrimprovefromraw} compared with the original models, indicating {\tool}'s effectiveness in improving driving model performance.
A qualitative study using traffic datasets from {\numdataset} geographically diverse regions shows that {\tool}-generated scenarios align closely with the corresponding traffic distributions in these regions. Looking forward, {\tool} can facilitate the automatic construction of traffic-safety scenario benchmarks, enabling consistent testing and targeted improvement for autonomous driving models.

%% file: sec/07_acknowledgement.tex
\section*{Acknowledgement}
We would like to thank the anonymous reviewers for their valuable feedback. 
This work is partly
supported by the National Science Foundation under the grant IIS-2416835.

%% file: sec/X_suppl.tex
\setcounter{page}{1}
\maketitlesupplementary
\appendix
\setcounter{figure}{3}
\setcounter{table}{3}

\section{Qualitative Study on Traffic Scenario Distribution Alignment}
\label{supp:sec:qualitative}
In this section, we present qualitative results on all {\numdataset} geographically diverse regions considered in our study, complementing the subset of three regions reported in Section~4.4 due to page limit. 
For each region, we visualize the embedding-space distributions of traffic scenarios generated by {\tool} and the five unaligned LLM baselines, alongside real-world traffic scenarios.
\input{supp/supp_fig_vis_baselines}

Across all {\numdataset} regions, we observe the same qualitative trend as in the Section~4.4 Figure~3): {\tool}-generated scenarios consistently co-locate with and overlap the real-world cluster, indicating close alignment with real-world traffic distributions. In contrast, GPT-4o, Claude-Sonnet-4, DeepSeek-V3, Qwen3, and GPT-5 form clusters that are clearly displaced from the real-world region, with limited or no overlap. This consistent pattern across all datasets further supports that {\tool} captures the dominant modes of the real-world traffic distribution and mitigates distribution shift relative to unaligned LLM baselines.

\newpage
\section{Ablation Study on Time Window Selection}
{\tool} uniformly samples frames from videos every 15 seconds (FPS=1/15), since the videos we collected from YouTube mainly depict everyday driving scenarios and rarely include short-horizon edge cases, making 15 seconds sufficient.
However, in more intense driving videos, shorter time windows may better capture short-horizon dynamics.
To examine this, we curate 10 intense driving videos from YouTube and then compare 5s vs. 15s sampling rate under the same settings as Section~4.2 and 4.3.

\Cref{supp:tab-time-window} shows the effectiveness of the scenarios in testing and improving autonomous driving model performances.
The results show that a shorter time window can indeed catch more challenging driving scenarios and enhance the effectiveness of scenario-based testing.
An interesting future direction is to adaptively detect short-horizon interactions using keyframe sampling methods instead of suing s fixed sampling interval.
\input{supp/supp_tab_time_window}

\newpage

\newpage
\section{Additional Details on Data Synthesis}
\subsection{Prompt for Traffic Scenario Synthesis}
The prompt used to synthesize traffic scenarios from real-world driving videos with GPT 4.1 nano (Section~3.1) is shown in the following box.
\input{supp/supp_prompt_extraction}

\clearpage
\subsection{Natural Language Description Examples}
This section provides two example natural language scenario descriptions (Section~3.1).
\input{supp/supp_nl_description_eg1}
\input{supp/supp_nl_description_eg2}

\clearpage
\section{Additional Details on Data Validation}

\subsection{Details of the Domain-Specific Language}
For efficient data validation (Section~3.2), we adopt the existing domain-specific language (DSL) design proposed by TARGET~\cite{DBLP:journals/tse/DengTYZZZ25}.
Building on the original grammar, we extend the DSL with an \texttt{actor\_group} element that groups actors located in close proximity and a \texttt{lane\_index} attribute that specifies lane-level positions.
These extensions enable a concise and precise representation of scenes with many participants.
As defined in Figure~\ref{fig:dsl}, the DSL follows a context-free grammar, comprising three sections: \emph{Environment}, \emph{Road Network}, and \emph{Actors}.
\begin{itemize}
    \item \textbf{\emph{Environment}} encodes time of day and weather conditions. 
    \item \textbf{\emph{Road network}} encodes road type, lane configurations, and traffic signal settings.
    \item \textbf{\emph{Actors}} encode the number, type, position, lane occupancy, and behaviors (e.g., go straight, turn left, etc.) of vehicles and pedestrians grouped by positions. 
\end{itemize}
\input{supp/fig_dsl}

\clearpage
\subsection{Prompt for Natural Language to DSL Translation}
The prompt used to translate {\tool}-synthesized traffic scenario natural language descriptions to the driving videos with GPT 5 (Section 3.1) is shown in the following box.
\input{supp/supp_prompt_description_to_dsl}

\newpage
\subsection{Effectiveness of DSL Validation}
We conduct a preliminary comparison to compare driving model testing and fine-tuning performance on scenarios synthesized with and without the DSL component.

As shown in \Cref{supp:tab-dsl-validation}, incorporating DSL validation is associated with improves scenario quality across both evaluation settings. 
For autonomous driving model testing experiments, scenarios generated with DSL validation achieve a lower overall score (0.319 vs. 0.408) and a higher collision rate (0.925 vs. 0.873). 
A similar trend is also observed when using the generated scenarios in fine-tuning autonomous models, where DSL validation improves the overall score from 0.369 to 0.402 and lowers the collision rate from 0.745 to 0.691. 
\input{supp/supp_tab_dsl_validation}

\vspace{2em}
\subsection{Preliminary Evaluation of DSL Validation Accuracy}
We conduct a preliminary evaluation to assess the accuracy of DSL validation and to examine the extent to which its outputs are affected by LLM hallucinations.
Specifically, we sample 120 natural language scenario descriptions, manually annotate the ground-truth DSL, and compare the DSL outputs generated by the DSL converter against these annotations.
We find that 13 out of 120 files 
contain hallucinations in the generated DSL, indicating that the DSL converter is affected by LLM hallucinations. Nevertheless, the error rate is relatively limited, and the hallucinations are concentrated in a small number of fields.
The most frequent hallucination is \texttt{lane\_number}, which appears in 9 files, followed by \texttt{one-way or two-way}, which appears in 4 files. We also observe one file each with a hallucination in \texttt{lane\_index} and the actor \texttt{type}.

Overall, these results indicate that hallucinations are present but not pervasive in DSL validation on this preliminary benchmark.

\clearpage
\section{Additional Details on LLM Alignment}

\subsection{Implementation Details for LLM Alignment}
\label{sec:supp-align-details}
We implement the fine-tuning pipeline using the Unsloth framework to leverage 4-bit quantization, enabling memory-efficient training of the Llama-3.2-3B-Instruct model on a single NVIDIA Tesla T4 GPU. 
The model is trained with a maximum sequence length of $2,048$ tokens. 
For Parameter-Efficient Fine-Tuning (PEFT), we configure Low-Rank Adaptation (LoRA) with a rank $r=16$ and a scaling factor $\alpha=16$. 
We apply LoRA adapters to all linear projection layers, including the query (\texttt{q\_proj}), key (\texttt{k\_proj}), value (\texttt{v\_proj}), output (\texttt{o\_proj}), and the MLP layers (\texttt{gate\_proj}, \texttt{up\_proj}, \texttt{down\_proj}), with no dropout. 
Optimization is performed using the 8-bit AdamW optimizer with a weight decay of $0.01$. 
We utilize a linear learning rate scheduler with $5$ warmup steps, targeting a peak learning rate of $2e-4$. 
The training process uses a per-device batch size of $2$ with $4$ gradient accumulation steps. 
To ensure the model learns strictly from the desired output distributions, we employ a response-only masking strategy, where loss is calculated exclusively on the assistant's generated tokens, masking the system instructions and user prompts.

\vspace{3ex}
\subsection{Overview of the Six LLM Alignment Datasets}
Table~\ref{supp:tab:datasets} summarizes the composition of the six LLM alignment datasets used in our study across diverse geographic locations.
\input{supp/tab_datasets.tex}

Across the six alignment datasets, the average number of actors clearly tracks the urban–rural divide. The three urban datasets, Los Angeles, New York City, and Switzerland, consistently exhibit dense interactions, with 13.8, 13.3, and 11.1 actors per scenario, respectively. In contrast, the rural and small-town datasets (Yellowstone, Yosemite, and Small Towns in PA) are much sparser, with only 2.3--3.2 actors on average.

\newpage
\subsection{Behavior Difference Across Geographic Regions}
We analyze generated scenarios from a metropolitan (New York City) and a rural (Yellowstone) dataset. 
As shown in \Cref{supp:fig-dataset-diff}, New York City is more crowded, with frequent pedestrian crossings and more parked sedans, while Yellowstone is sparser and features rural-only actors (e.g., deer, cows, bison, tractors) absent in New York City.
\input{supp/supp_fig_dataset_diff}

\vspace{1em}
\subsection{Human Study on Scenario Semantic Fidelity}
To evaluate whether {\tool}-generated scenarios are semantically consistent with real-world traffic scenarios, we conduct a qualitative study in Section 4.4 and \Cref{supp:sec:qualitative}.
In this section, we further conduct a human study to assess the semantic fidelity of the generated scenarios. 
We sample 100 scenarios (50 New York City, 50 Yellowstone) and ask 6 graduate students to independently rate the semantic fidelity of each scenario on a 1–5 scale, measuring “how likely the traffic scenario is to occur in the specified location”. 
The human study reports mean scores of {4.553} (New York City) and {4.627} (Yellowstone).
The results show that {\tool} generates semantically faithful, region-consistent traffic scenarios.

\vspace{3ex}
\subsection{{\textbf{\tool}} Prompt for Real-World Aligned Traffic Scenario Generation}
This section presents the complete prompt used in our experiments to query the aligned LLMs introduced in Section~3.3. 
\input{supp/supp_prompt_ours_aligned}

\section{Unaligned LLM Baseline Prompts}
The prompt used to generate traffic scenarios using unaligned LLM baselines in the ablation study in Section~4.1 is shown in the following box. 
\input{supp/supp_prompt_LLMbaselines}

%% file: supp/supp_fig_vis_baselines.tex
\begin{figure*}[htb]
  \centering
  \begin{subfigure}{0.32\textwidth}
    \includegraphics[width=\linewidth]{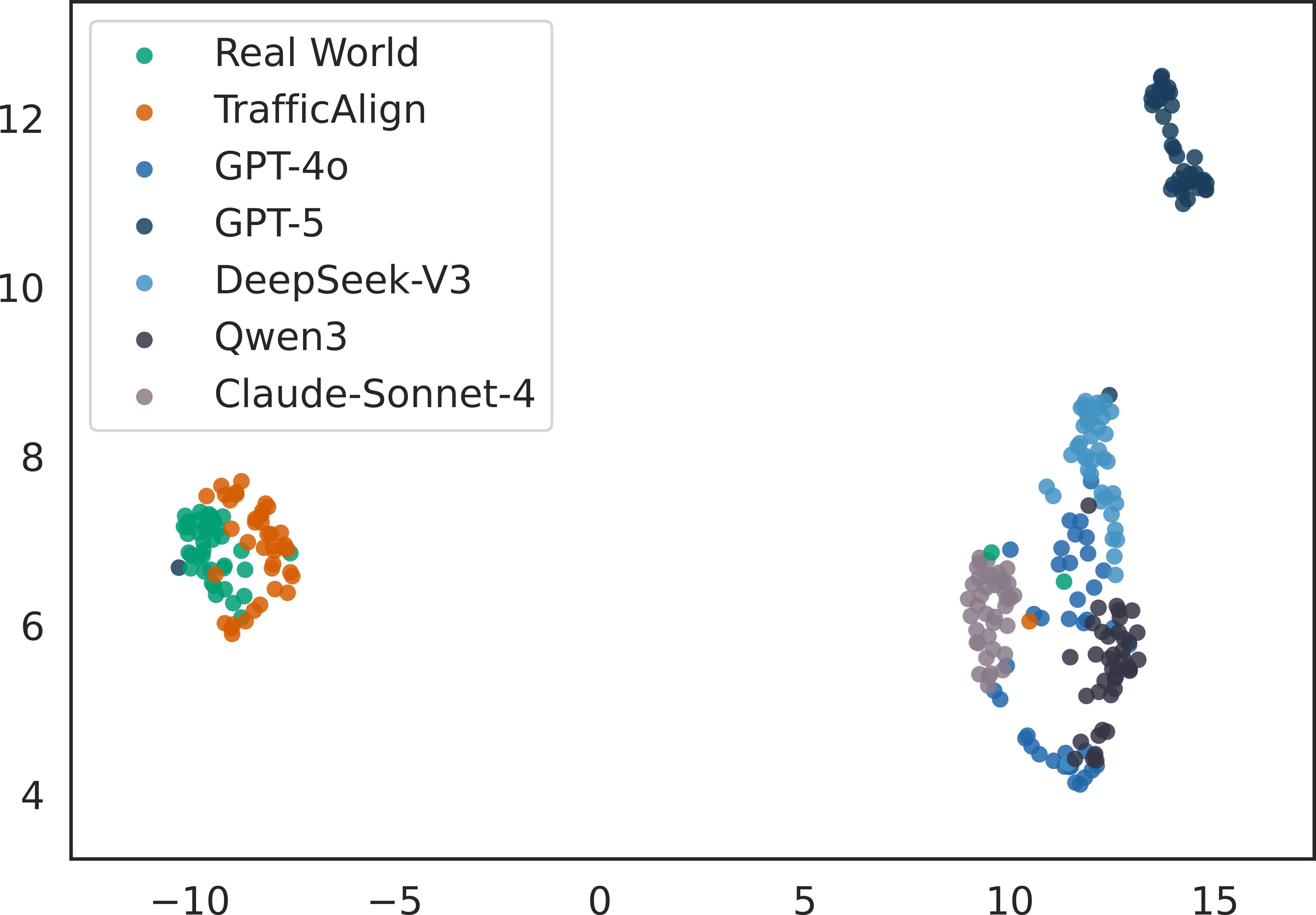}
    \caption{Los Angeles.}
    \label{fig:supp-vis-baselines-la}
  \end{subfigure}
  \hfill
  \begin{subfigure}{0.32\textwidth}
    \includegraphics[width=\linewidth]{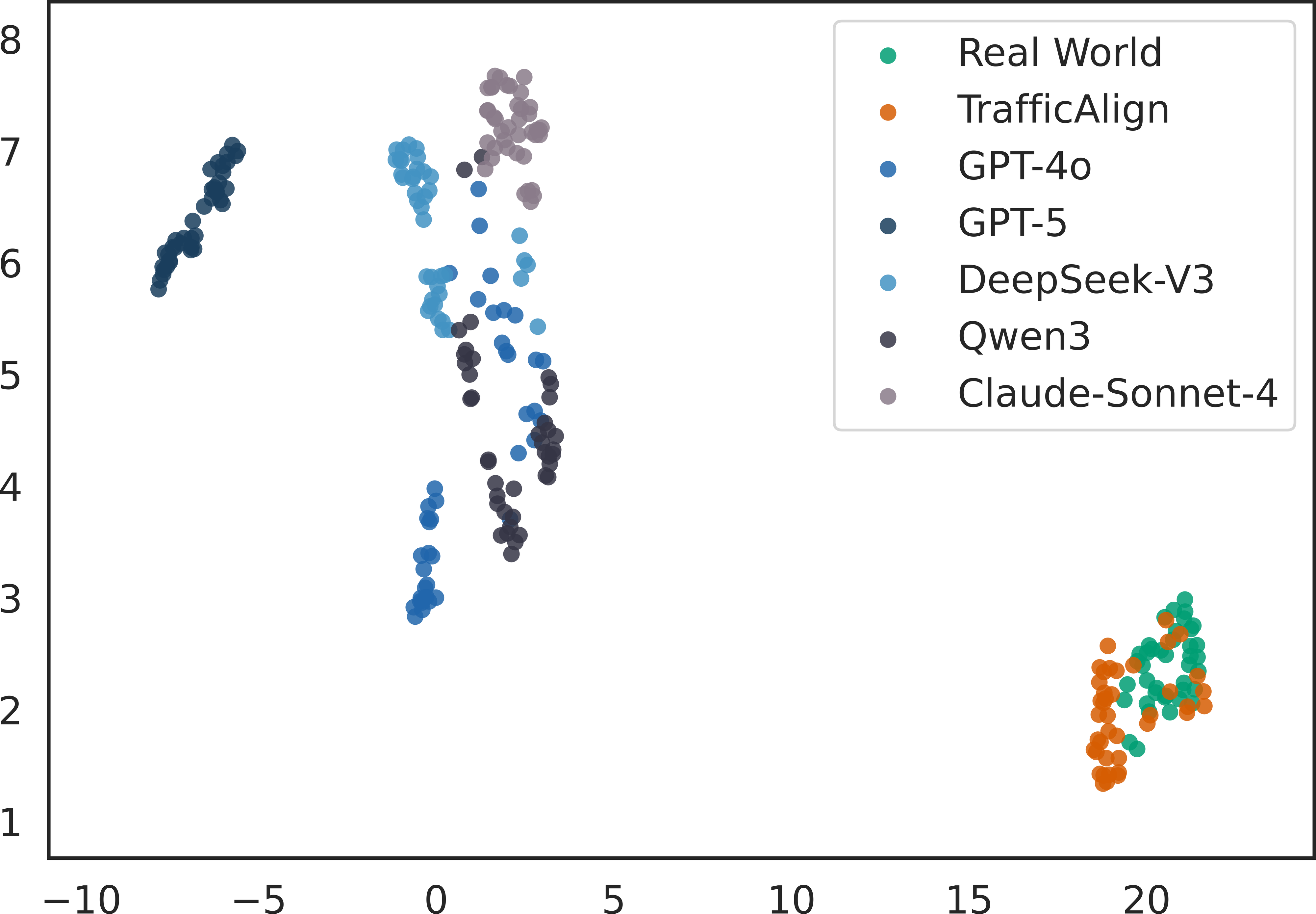}
    \caption{Yellowstone.}
    \label{fig:supp-vis-baselines-yellowstone}
  \end{subfigure}
  \hfill
  \begin{subfigure}{0.32\textwidth}
    \includegraphics[width=\linewidth]{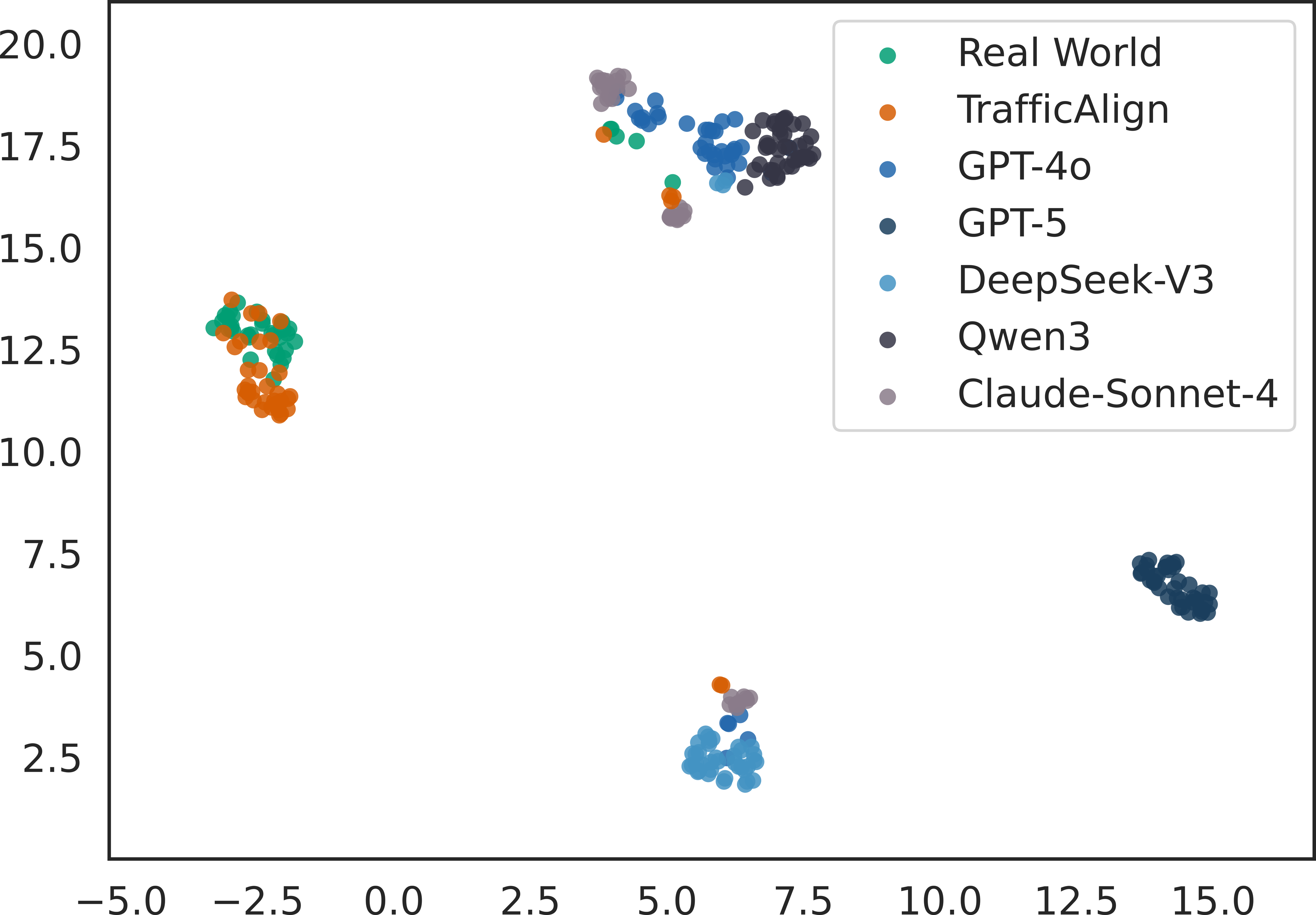}
    \caption{Small Towns in Pennsylvania, USA.}
    \label{fig:supp-vis-baselines-smalltowns}
  \end{subfigure}
  
  \vspace{3ex}
  
  \begin{subfigure}{0.32\textwidth}
    \includegraphics[width=\linewidth]{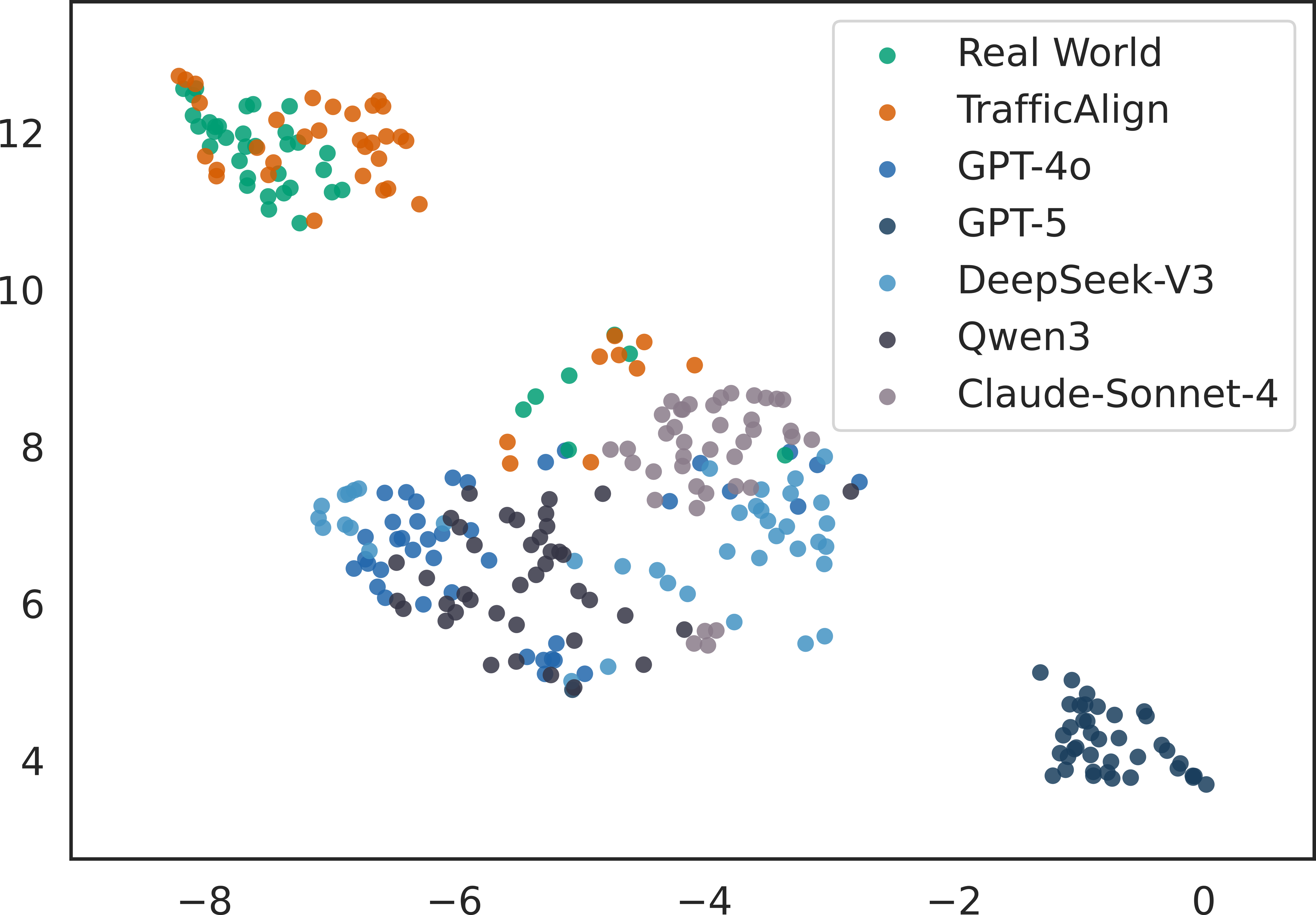}
    \caption{New York City.}
    \label{fig:supp-vis-baselines-nyc}
  \end{subfigure}
  \hfill
  \begin{subfigure}{0.32\textwidth}
    \includegraphics[width=\linewidth]{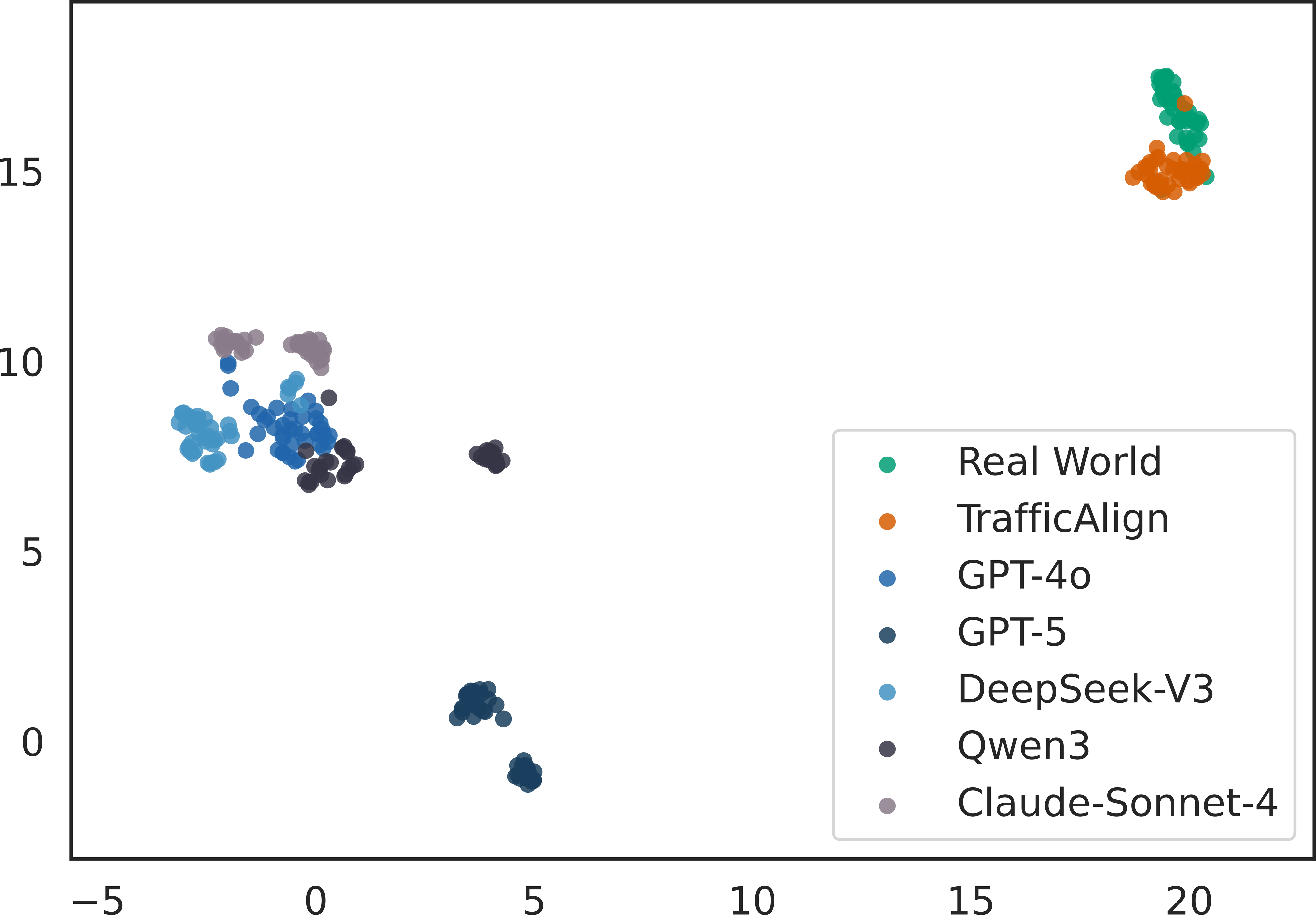}
    \caption{Yosemite.}
    \label{fig:supp-vis-baselines-yosemite}
  \end{subfigure}
  \hfill
  \begin{subfigure}{0.32\textwidth}
    \includegraphics[width=\linewidth]{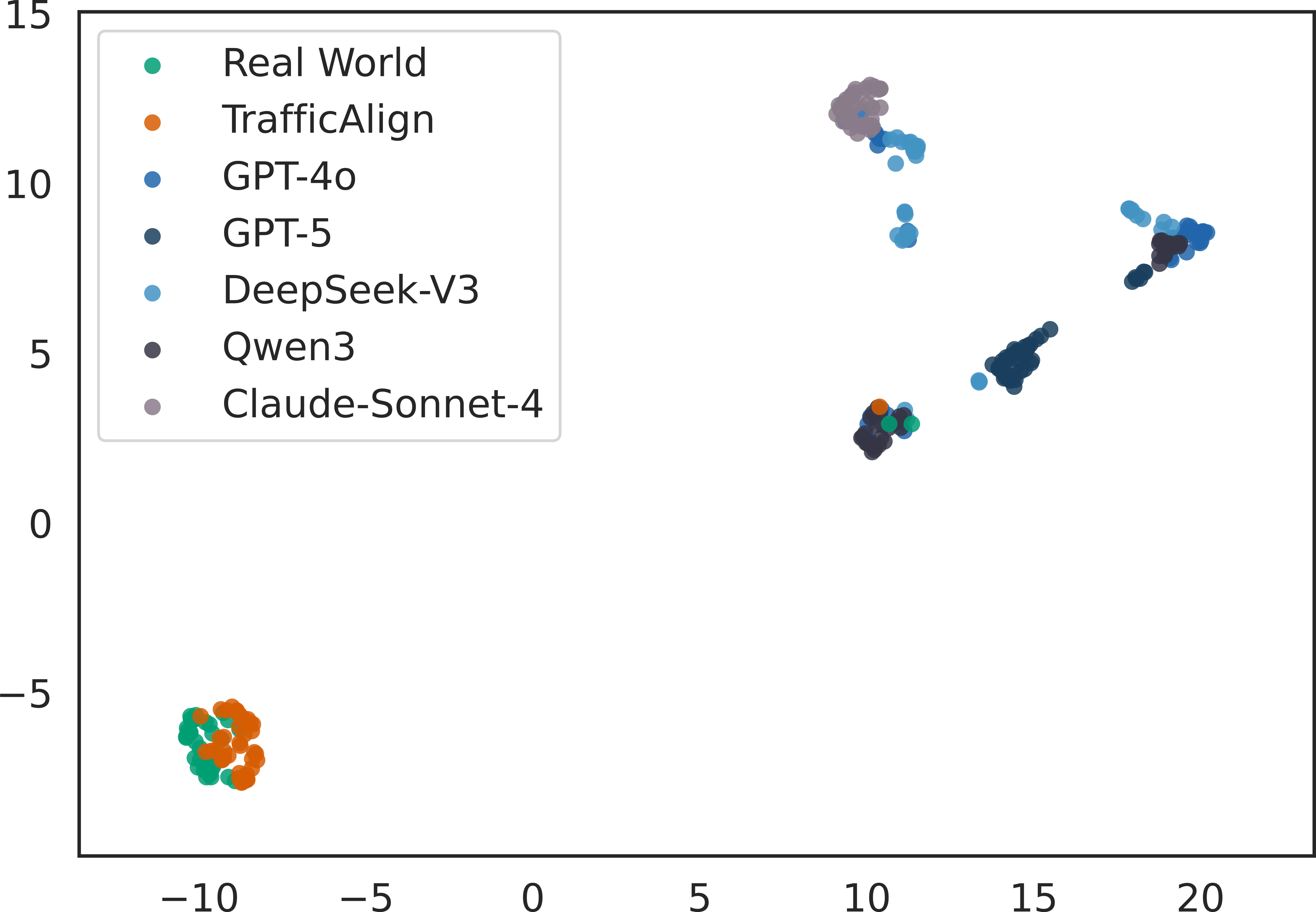}
    \caption{Switzerland.}
    \label{fig:supp-vis-baselines-switzerland}
  \end{subfigure}

\caption{UMAP visualization of embeddings of traffic scenarios generated by {\tool}, {\numllmbaseline} unaligned LLM baselines, and real-world traffic scenarios.}
\label{fig:supp-vis-baselines}
\end{figure*}

%% file: supp/supp_tab_time_window.tex
\begin{table}[htb]
\centering
\caption{
\textbf{Scenario effectiveness across time windows.}
This table presents the average test results on {\numads} distinct autonomous driving models using scenarios generated from ten intense videos
by {\tool} using frame sampling rates of 5 seconds and 15 seconds. 
CR: collision rate, RR: frequency of running red lights, SS: frequency of running stop signs, OR: average distance driven out of road, RF: route following stability, Comp: average percentage of route completion, TS: average time spent to complete the route, ACC: average acceleration, YV: average yaw velocity, LI: frequency of lane invasion, OS: overall score.
$\uparrow$/$\downarrow$: higher/lower values indicate more effective in testing the driving models.
\textbf{Values in bold} indicate the best performance; 
}
\label{supp:tab-time-window}
\centering
\begin{subtable}[t]{0.75\linewidth}
\centering
\caption{Testing driving models.}
\resizebox{\linewidth}{!}{%

\begin{tabular}{l|cccc|ccc|ccc|c}
\toprule
\multicolumn{1}{c|}{\multirow{2}{*}{\textbf{Method}}} & \multicolumn{4}{c|}{\textbf{Safety Level}} & \multicolumn{3}{c|}{\textbf{Functionality Level}} & \multicolumn{3}{c|}{\textbf{Etiquette Level}} & \multirow{2}{*}{\textbf{OS $\downarrow$}} \\
\multicolumn{1}{c|}{} & CR $\uparrow$ & RR $\uparrow$ & SS $\uparrow$ & OR $\uparrow$ & RF $\downarrow$ & Comp $\downarrow$ & TS $\uparrow$ & ACC $\uparrow$ & YV $\uparrow$ & LI $\uparrow$ &  \\ \midrule
{\tool} (5s) & \textbf{0.907} & \textbf{0.469} & 0.183 & \textbf{0.207} & \textbf{0.741} & 0.582 & \textbf{0.477} & 0.756 & \textbf{0.660} & 0.218 & \textbf{0.431} \\
{\tool} (15s) & 0.899 & 0.456 & \textbf{0.221} & 0.204 & 0.785 & \textbf{0.438} & 0.298 & \textbf{0.809} & 0.476 & \textbf{0.355} & 0.441 \\ \bottomrule
\end{tabular}%

}
\end{subtable}
\vfill
\centering
\begin{subtable}[t]{0.75\linewidth}
\vspace{1em}
\centering
\caption{Fine-tuning driving models.}

\resizebox{\linewidth}{!}{%
\begin{tabular}{l|cccc|ccc|ccc|c}
\toprule
\multicolumn{1}{c|}{\multirow{2}{*}{\textbf{Method}}} & \multicolumn{4}{c|}{\textbf{Safety Level}} & \multicolumn{3}{c|}{\textbf{Functionality Level}} & \multicolumn{3}{c|}{\textbf{Etiquette Level}} & \multirow{2}{*}{\textbf{OS $\uparrow$}} \\
\multicolumn{1}{c|}{} & CR $\downarrow$ & RR $\downarrow$ & SS $\downarrow$ & OR $\downarrow$ & RF $\uparrow$ & Comp $\uparrow$ & TS $\downarrow$ & ACC $\downarrow$ & YV $\downarrow$ & LI $\downarrow$ &  \\ \midrule
{\tool} (5s) & \textbf{0.766} & \textbf{0.471} & 0.007 & 0.005 & \textbf{0.826} & \textbf{0.528} & 0.311 & 0.631 & 0.533 & \textbf{0.513} & \textbf{0.396} \\
{\tool} (15s) & 0.768 & 0.476 & \textbf{0.000} & \textbf{0.003} & 0.813 & 0.496 & \textbf{0.298} & \textbf{0.629} & \textbf{0.519} & 0.524 & 0.391 \\ \bottomrule
\end{tabular}%
}

\end{subtable}
\end{table}

%% file: supp/supp_prompt_extraction.tex
\begin{tcolorbox}[
    breakable,
    enhanced jigsaw, 
    width=\textwidth,
    colframe=black,
    colback=white,
    boxrule=1pt,
    arc=2mm,
    title={\textbf{Prompt for Traffic Scenario Synthesis from Real-World Driving Video Frames}},
    fonttitle=\bfseries\large,
    coltitle=white,
    colbacktitle=black,
    center title,
    bottom=1mm,
    top=1mm,
    boxsep=3mm,
    before skip=10pt,
    after skip=15pt,
]
\label{fig:supp-prompt-extract}
\textbf{\color{slategray}SYSTEM}

You are a traffic domain expert. Analyze the given image and extract structured information about the traffic scenario.

\vspace{2ex}
\textbf{\color{slategray}TASK}

Analyze the given image and extract information about the traffic scenario following the steps below.
Approach this task step-by-step, take your time, and don't skip steps.

Extract the basic information of the traffic scenario:
\begin{enumerate}[label={Step \arabic*.}]
    \item
    \texttt{weather}: one of [\textit{clear, cloudy, foggy, rainy, snowy}], or describe in your own words.

    \item
    \texttt{time}: one of [\textit{daytime, nighttime}], or describe in your own words.

    \item
    \texttt{road\_type}: one of [\textit{straight road, intersection, t-intersection, roundabout}].

    \item
    \texttt{one-way\_or\_two-way}: Determine carefully from visual evidence. A road is one-way if: (a) all visible moving vehicles travel in the same direction, (b) lane markings show no center divider separating opposing flows, (c) one-way signs are visible, or (d) there is no opposing traffic lane. A road is two-way \emph{only} if opposing traffic lanes or oncoming vehicles are clearly visible. Do NOT default to two-way; state one-way whenever opposing traffic is absent. You MUST end this step with an explicit conclusion: state exactly \textit{``This is a one-way road.''} or \textit{``This is a two-way road.''} with no other phrasing.

    \item
    \texttt{special\_lane}: note any special lanes (e.g., bike lane, tram track, bus lane), or state \texttt{none}.

    \item
    \texttt{lane\_number}: Count only the lanes visible in the ego vehicle's direction of travel. Use a precise integer. Estimate if unknown.

    \item
    \texttt{traffic\_sign}: one of [\textit{stop sign, speed limit sign, yield sign, yield to pedestrian sign, school zone}], or state \texttt{none}.

    \item
    \texttt{traffic\_light}: one of [\textit{green, red, yellow, broken}], or state \texttt{none}.

    \item
    \texttt{road\_context}: one of [\textit{urban, residential, highway, country road, dirt road, hill road}], or describe in your own words.
\end{enumerate}
\begin{enumerate}[label={Step \arabic*.}, start=10]
    \item
    Observe additional and detailed information of the traffic scenario:
    \begin{itemize}
        \item Whether the traffic is heavy or light. Is there a traffic jam?
        \item Whether there are any traffic accidents or other road hazards.
        \item Whether there are any road works or lane closures.
        \item Whether there are any emergency vehicles.
        \item Describe the roadside environment, including buildings, trees, and other objects.
    \end{itemize}

    \item
    Observe any other information that you think is important.

    \item
    \texttt{actors}: The input image is captured by a dashboard (forward-facing) camera mounted on the ego vehicle. This means ONLY actors in FRONT of or BESIDE the ego vehicle are visible. Do NOT describe any actor as being behind, left behind, or right behind the ego vehicle.

    Before listing actors, scan the full image systematically: near field (close to the ego vehicle), mid field, and far field; left side, center, and right side. Include ALL actors that are present in the image, even if they are partially visible or at the edge of the frame. Do NOT omit any actor that can be seen. At the same time, do NOT invent or assume actors that have no visual evidence in the image.

    For each detected actor (e.g., sedan, bus, pedestrian, bicycle, etc.), record the following attributes. Remember to also describe the ego vehicle.
    \begin{itemize}
        \item \texttt{type}: The classification of the actor (e.g., sedan, SUV, truck, pedestrian, bicycle, etc.).

        \item \texttt{current\_behavior}: The actor's present action or intent (e.g., move forward, turn left, stopped, yield).

        \item \texttt{speed}: The current speed of the actor as an integer in km/h. You MUST provide a numeric estimate --- never use vague descriptions such as ``slow'', ``fast'', ``moderate'', or ``unknown''. If the exact speed cannot be determined, estimate it based on context:
        \begin{itemize}
            \item stopped vehicle $\Rightarrow$ \texttt{0 km/h}
            \item city vehicle moving normally $\Rightarrow$ \texttt{30--50 km/h}
            \item highway vehicle $\Rightarrow$ \texttt{80--120 km/h}
        \end{itemize}

        \item \texttt{position\_target} and \texttt{position\_relation}: Relative position to the ego vehicle. The valid values and their precise definitions are:
        \begin{itemize}
            \item \texttt{front}: the actor is in the \emph{same lane} as the ego vehicle and is ahead.
            \item \texttt{left front}: the actor is in a lane to the \emph{left} of the ego vehicle AND is ahead longitudinally.
            \item \texttt{right front}: the actor is in a lane to the \emph{right} of the ego vehicle AND is ahead longitudinally.
            \item \texttt{left}: the actor is in a lane to the \emph{left} at roughly the same longitudinal position (side-by-side).
            \item \texttt{right}: the actor is in a lane to the \emph{right} at roughly the same longitudinal position (side-by-side).
        \end{itemize}
        The relations \texttt{behind}, \texttt{left behind}, and \texttt{right behind} must NOT be used. Left and right are always from the ego vehicle driver's perspective (same as left/right in the image). To assign the correct relation, first determine the actor's lane relative to the ego vehicle (same / left / right), then determine whether the actor is ahead or side-by-side.

        \item \texttt{lane\_index}: An integer for the lane within the road network. The leftmost lane in the ego vehicle's direction is index \texttt{1}; the leftmost lane in the opposite direction is \texttt{-1}.
        \begin{itemize}
            \item E.g., ``parked in the rightmost lane'' with \texttt{lane\_number: 3} $\Rightarrow$ \texttt{lane\_index: 3}.
            \item E.g., ``moving forward in the leftmost lane'' with \texttt{lane\_number: 2} $\Rightarrow$ \texttt{lane\_index: 1}.
        \end{itemize}
    \end{itemize}

    \item
    Generate a concise, technical description of the traffic scenario that:
    \begin{itemize}
        \item Uses precise traffic engineering terminology, active voice, and present tense.
        \item Uses assertive language (\textit{``is''/``are''} instead of \textit{``appears''/``seems''}).
        \item Avoids image-related terms (e.g., ``shows'', ``displays'', ``depicts'').
        \item Uses precise numbers (e.g., ``10 sedans'' instead of ``many'').
        \item Describes road network first (road context, road type, one-way/two-way, special lanes, lane number, traffic sign, traffic light), then environment (weather, time of day), then additional details from Steps 10--11.
        \item Ends with an \textbf{Actors} section (bold title on a new line) listing each actor type with its precise count.
    \end{itemize}

\end{enumerate}

\vspace{2ex}
\textbf{\color{slategray}OUTPUT FORMAT}

{\textless}ANALYSIS{\textgreater}

[Your chain of thoughts following the step-by-step instructions provided in step 1 to step 12]

{\textless}/ANALYSIS{\textgreater}

\vspace{1ex}
{\textless}TEXT{\textgreater}

[Insert paragraph description here]

**Actors**

[Insert paragraph describing the actors here]

{\textless}/TEXT{\textgreater}

\vspace{2ex}
\textbf{\color{slategray}IN CONTEXT LEARNING EXAMPLE}

\textbf{Example input image:}

\includegraphics[width=0.5\textwidth]{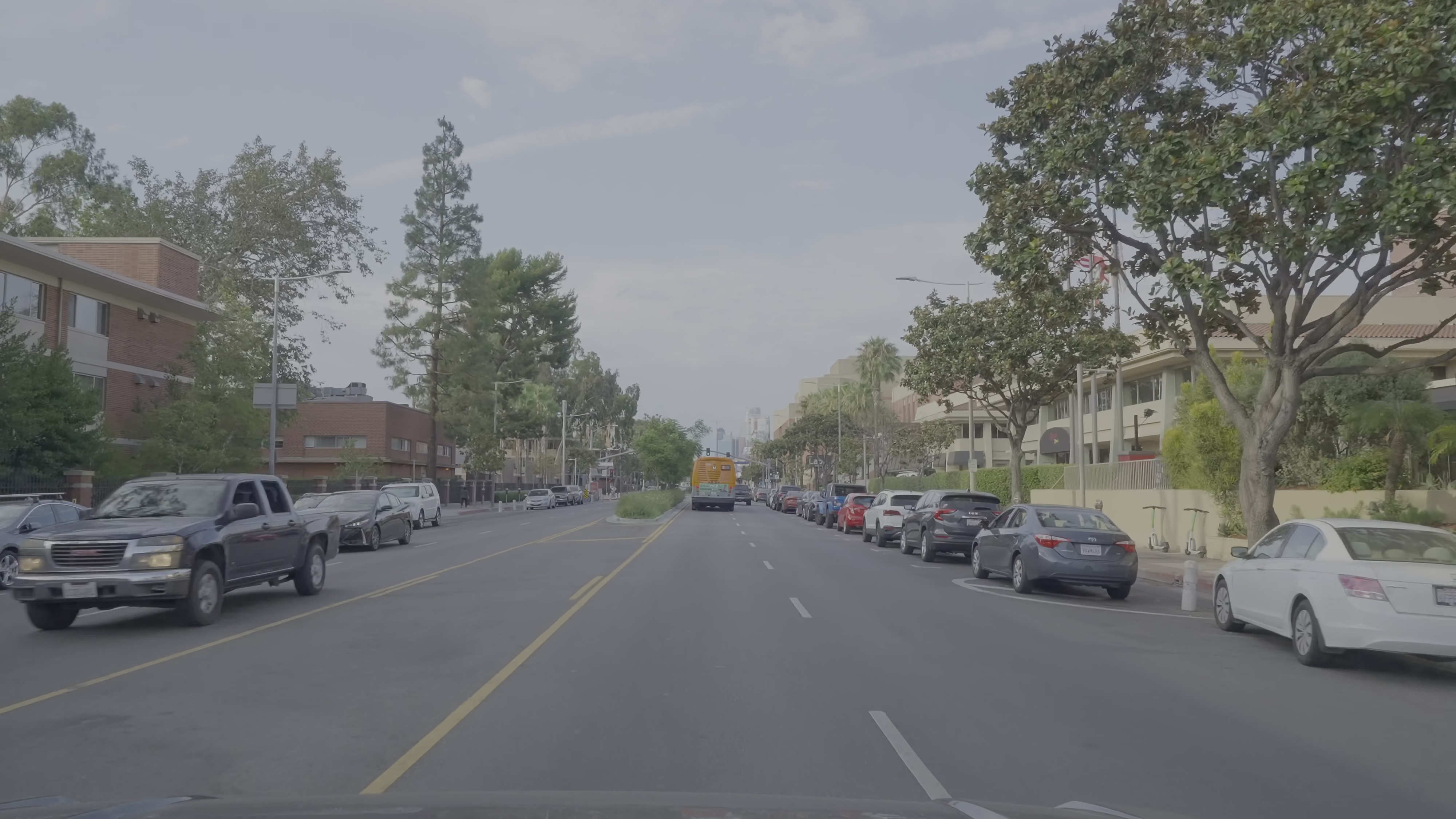}

\clearpage
\textbf{Example output:}

{\textless}ANALYSIS{\textgreater}
\begin{enumerate}[label={Step \arabic*.}]
    \item
    weather: clear
    \item
    time: daytime
    \item
    road type: straight road
    \item
    one-way or two-way: two-way
    \item
    special lane: None
    \item
    lane number: 6
    \item
    traffic sign: no traffic sign is observed
    \item
    traffic light: The traffic light is current green
    \item
    road context: urban, residential and commercial mixed-use area
    \item
    the traffic is light; there are no traffic accidents; there are no road works or lane closures; there are no special vehicles such as emergency vehicles
    \item
    No special components are worth noticing
    \item
    actors: Several types of actors are observed. The ego vehicle travels forward at about 38~km/h. A bus is in front of the ego vehicle moving at around 35~km/h. Approximately 10 sedans are parked along the street in the ego direction, and 7 sedans are parked on the rightmost lane in the opposite direction. A truck is driving toward the ego vehicle in the central oncoming lane at about 40~km/h.
\end{enumerate}
{\textless}/ANALYSIS{\textgreater}

\vspace{2ex}
{\textless}TEXT{\textgreater}

The traffic scenario occurs on an urban, two-way road with a standard lane configuration, three lanes per direction. 
The road is marked with one solid yellow centerline, indicating no overtaking, and has no speed limit sign present.
A traffic light is visible and currently displays green, enabling vehicle passage in the direction of travel. The environment is during daytime with clear weather and bright conditions. The road is bordered by residential and commercial buildings, with trees lining both sides of the street, and sidewalks are visible along the curb. No traffic jams or accidents are evident, and there are no ongoing road works or lane closures. No emergency vehicles are present. The roadside vegetation, including trees and shrubbery, appears well-maintained, and the overall environment is calm. Traffic flow is smooth with no signs of congestion or hazards, and pedestrians are not observed in the scene.

\textbf{**Actors**}

The ego vehicle is traveling forward at about 38 km/h. There is a bus driving in front of the ego vehicle at around 35 km/h. There are approximately 10 sedans parked along the sides of the street in the ego vehicle's direction and 7 parked sedans on the rightmost lane in the opposite direction. There is a truck driving in the opposite direction in the most central lane at about 40 km/h.

{\textless}/TEXT{\textgreater}

\vspace{2ex}
\textbf{Input image:}

[Insert the input image here.]

\end{tcolorbox}

%% file: supp/supp_nl_description_eg1.tex
\begin{tcolorbox}[
    breakable,
    enhanced jigsaw, 
    width=\textwidth,
    colframe=black,
    colback=white,
    boxrule=1pt,
    arc=2mm,
    title={\textbf{Natural Language Scenario Description Example 1}},
    fonttitle=\bfseries\large,
    coltitle=white,
    colbacktitle=black,
    center title,
    bottom=1mm,
    top=1mm,
    boxsep=3mm,
    before skip=10pt,
    after skip=15pt,
]

The scenario unfolds on a straight, two-lane urban roadway with a single direction of travel and no special-purpose lanes. Conditions are daytime with clear skies, providing good visibility. The surroundings consist of commercial buildings, sidewalks, and typical city infrastructure, including several tall structures. Traffic signals are green, permitting continuous vehicle flow, and there are no signs of congestion, incidents, or construction work. The road surface is unobstructed, contributing to smooth travel conditions. No emergency vehicles or hazardous elements are detected, and lane availability is unrestricted, indicating all lanes remain open for normal use.

\textbf{**Actors**}

The ego vehicle occupies the right lane, proceeding forward at 30 km/h. In the left lane, a bus travels straight at roughly 25 km/h. Two additional vehicles are present: a car ahead of the ego vehicle and an SUV positioned directly ahead of the bus, each moving at about 30 km/h.

\end{tcolorbox}

%% file: supp/supp_nl_description_eg2.tex
\begin{tcolorbox}[
    breakable,
    enhanced jigsaw, 
    width=\textwidth,
    colframe=black,
    colback=white,
    boxrule=1pt,
    arc=2mm,
    title={\textbf{Natural Language Scenario Description Example 2}},
    fonttitle=\bfseries\large,
    coltitle=white,
    colbacktitle=black,
    center title,
    bottom=1mm,
    top=1mm,
    boxsep=3mm,
    before skip=10pt,
    after skip=15pt,
]

The scene depicts a four-lane, straight urban roadway with single-direction traffic during bright daylight with clear weather beneath an overcast sky. Trees and high-rise buildings line both sides of the street. No traffic lights regulate movement. 
At the approaching intersection, a ``Stop'' sign controls vehicle behavior. 
The roadside environment consists of sidewalks, trees, and dense urban structures. 
Traffic is partially free-flowing with minimal congestion or hazards, and several vehicles are already in motion. 

\textbf{**Actors**}

The ego vehicle is positioned in the second lane from the right, traveling forward at approximately 35 km/h. Seven pedestrians are crossing at the marked crosswalk. Two taxis are moving in the outer-right lane at roughly 30 km/h, while around ten stationary vehicles are parked or stopped along the right roadside. An unoccupied bus is positioned along the right-side curb, remaining stationary throughout the scene.
\end{tcolorbox}

%% file: supp/fig_dsl.tex
\begin{figure}[htb]
\centering
\resizebox{0.5\linewidth}{!}{%
$\def\arraystretch{1.1}
\setlength{\arraycolsep}{1pt}
\begin{array}{lll}
\textbf{\emph{Scenario}} & ::= & \emph{Environment} ; \emph{Road\_network} ; \emph{Actors} \\
\textbf{\emph{Environment}} & ::= & \emph{weather}; \ \emph{time} \\
\emph{weather} & ::= & \texttt{rainy} \ | \ \texttt{foggy} \ | \ \texttt{snowy} \ | \ \texttt{wet} \ | \ \texttt{...} \\
\emph{time} & ::= & \texttt{daytime} \ | \ \texttt{nighttime} \\
\textbf{\emph{Road\_network}} & ::= & \emph{road\_type}; \ \emph{traffic\_signals}; \ \emph{lane\_number} \\
\emph{road\_type} & ::= & \texttt{intersection} \ | \ \texttt{roundabout} \ | \ \texttt{...} \\
\emph{traffic\_signals} & ::= & \emph{traffic\_signs}, \emph{traffic\_light} \\
\emph{traffic\_signs} & ::= & \epsilon \ | \ \emph{traffic\_sign}; \ \emph{traffic\_signs} \\
\emph{traffic\_sign} & ::= & \texttt{stop\_sign} \ | \ \texttt{speed\_limit\_sign} \ | \ \texttt{...} \\
\emph{traffic\_light} & ::= & \epsilon \ | \texttt{red\_light} \ | \ \texttt{green\_light} \\
\emph{lane\_number} & ::= & \texttt{0} \ | \ \texttt{1} \ | \ \texttt{2} \ | \ \texttt{3} \ | \ \texttt{...} \\
\textbf{\emph{Actors}} & ::= & \emph{ego\_vehicle}; \ \emph{npc\_actors} \\
\emph{ego\_vehicle} & ::= & \emph{behavior};  \ \emph{position}\\
\emph{npc\_actors} & ::= & \epsilon \ | \ \emph{actor\_group}; \ \emph{npc\_actors} \\
\emph{actor\_group} & ::= & \emph{actor\_number}; \ \emph{actor\_type}; \ \emph{behavior}; \ \emph{position} \\
\emph{actor\_type} & ::= & \texttt{car} \ | \ \texttt{truck} \ | \ \texttt{pedestrian} \ | \ \texttt{...} \\
\emph{behavior} & ::= & \texttt{go\_forward} \ | \ \texttt{turn\_left} \ | \ \texttt{...} \\
\emph{position} & ::= & \emph{reference\_point}; \ \emph{relative\_position}; \ \emph{lane\_index} \\
\emph{reference\_point} & ::= & \texttt{ego\_vehicle} \ | \ \emph{road\_type} \ | \ \emph{traffic\_signals} \\
\emph{relative\_position} & ::= & \texttt{front} \ | \ \texttt{behind} \ | \ \texttt{left} \ | \ \texttt{on} \ | \ \texttt{...} \\
\emph{lane\_index} & ::= & \texttt{0} \ | \ \texttt{1} \ | \ \texttt{2} \ | \ \texttt{3} \ | \ \texttt{...} \\
\end{array}$

}

\caption{The grammar of the traffic scenario DSL~\cite{DBLP:journals/tse/DengTYZZZ25}.}
\label{fig:dsl}
\end{figure}

%% file: supp/supp_prompt_description_to_dsl.tex
\begin{tcolorbox}[
    breakable,
    enhanced jigsaw,
    width=\textwidth,
    colframe=black,
    colback=white,
    boxrule=1pt,
    arc=2mm,
    title={\textbf{Prompt for Translating Natural Language Description into DSL Representation}},
    fonttitle=\bfseries\large,
    coltitle=white,
    colbacktitle=black,
    center title,
    bottom=1mm,
    top=1mm,
    boxsep=3mm,
    before skip=10pt,
    after skip=15pt,
]

\textbf{\color{slategray}SYSTEM}

You are a traffic domain expert familiar with domain-specific languages (DSL).

\vspace{2ex}
\textbf{\color{slategray}TASK}

Your task is to extract the information about a traffic scenario from a textual description and represent it as a DSL in YAML format.
Take your time to solve this task step by step.

\vspace{1ex}
\textbf{Steps 1--8:} Extract the basic information of the traffic scenario:

\begin{enumerate}[label={Step \arabic*.}]
    \item
    \texttt{weather}: one of [\textit{clear, cloudy, foggy, rainy, snowy}]. Don't infer. If the natural language description does not clearly state this, use \texttt{unspecified}.

    \item
    \texttt{time}: one of [\textit{daytime, nighttime}]. Don't infer. If the natural language description does not clearly state this, use \texttt{unspecified}.

    \item
    \texttt{road\_type}: one of [\textit{straight road, curved road, intersection, t-intersection, roundabout}]; include one-way/two-way and any special lanes (e.g., bike lane, tram track). Don't infer. If the natural language description does not clearly state this, use \texttt{unspecified}.

    \item
    \texttt{lane\_number}: The number of lanes \emph{per direction} for a two-way road. Use a precise integer. Don't infer. If not clearly stated, use \texttt{unspecified}.
    \begin{itemize}
        \item E.g., a two-way road with two lanes in each direction $\Rightarrow$ \texttt{lane\_number: 2}.
        \item E.g., a two-way road with two lanes total $\Rightarrow$ \texttt{lane\_number: 1}.
    \end{itemize}

    \item
    \texttt{one-way\_or\_two-way}: one of [\textit{one-way, two-way}]. Don't infer. If not clearly stated, use \texttt{unspecified}.

    \item
    \texttt{traffic\_sign}: one of [\textit{stop sign, speed limit sign, yield sign, yield to pedestrian sign, school zone}], or \texttt{none} if no sign is present. Don't infer. If not clearly stated, use \texttt{unspecified}.

    \item
    \texttt{traffic\_light}: one of [\textit{green, red, yellow, broken}], or \texttt{none} if no traffic light is present. Don't infer. If not clearly stated, use \texttt{unspecified}.

    \item
    \texttt{road\_context}: one of [\textit{urban, residential, highway, country road, dirt road, hill road}]. Don't infer. If not clearly stated, use \texttt{unspecified}.
\end{enumerate}

\vspace{1ex}
\begin{enumerate}[label={Step \arabic*.}, start=9]
    \item
    \texttt{actors}: For each actor group (e.g., car, bus, pedestrian, bicycle), extract the following attributes:
    \begin{itemize}
        \item \texttt{type}: The classification of the actor (e.g., sedan, SUV, truck, pedestrian, cyclist). Don't infer. If not clearly stated, use \texttt{unspecified}.

        \item \texttt{number}: The number of actors in this group. If the description states an approximate count using hedging words (e.g., ``around 10'', ``approximately 5'', ``about 3'', ``roughly 8''), extract the integer value --- do \emph{not} treat hedging language as \texttt{unspecified}. Only use \texttt{unspecified} when no numeric information is given at all (e.g., ``several'', ``a few'', ``many'').

        \item \texttt{behavior}: The actor’s present action or intent (e.g., move forward, turn left, stop, yield). Don’t infer. If the natural language description does not clearly state this, use ``unspecified''. Use ``parked'' when the actor is described as parked (e.g., ``parked along the roadside'', ``parked by the curb'') — do not use ``stopped'' for parked actors. Use ``stopped'' only for actors that have temporarily halted while in traffic (e.g., waiting at a red light). Treat ``stationary or moving slowly'' as \texttt{stopped}.

        \item \texttt{speed}: The current speed in km/h. If an approximate speed is given using hedging words (e.g., ``around 30 km/h'', ``approximately 50 km/h'', ``roughly 60 km/h''), extract the integer value. Use \texttt{unspecified} when: (a) no numeric speed information is given, or (b) speed is described only with a qualitative adjective (e.g., ``slow'', ``fast'', ``moderate'', ``crawling'') with no accompanying number.

        \item \texttt{position\_target} and \texttt{position\_relation}: Relative position to the ego vehicle. If impossible to infer, use \texttt{unspecified}.
        \begin{itemize}
            \item E.g., ``directly ahead of the ego vehicle in the same lane'' $\Rightarrow$ \texttt{position\_target: ego\_vehicle}; \texttt{position\_relation: front}.
            \item E.g., ``to the left rear of the ego vehicle in an adjacent lane'' $\Rightarrow$ \texttt{position\_target: ego\_vehicle}; \texttt{position\_relation: left behind}.
            \item E.g., ``in the opposite direction driving towards the ego vehicle'' $\Rightarrow$ \texttt{position\_target: ego\_vehicle}; \texttt{position\_relation: front}.
            \item E.g., ego vehicle is in the leftmost lane and another actor is parked by the curbside $\Rightarrow$ \texttt{position\_target: ego\_vehicle}; \texttt{position\_relation: right}.
        \end{itemize}

        \item \texttt{lane\_index}: An integer for the lane within the road network. The leftmost lane in the ego vehicle's direction is index \texttt{1}; the leftmost lane in the opposite direction is \texttt{-1}. Use \texttt{unspecified} if not determinable.
        \begin{itemize}
            \item E.g., ``parked in the rightmost lane'' with \texttt{lane\_number: 3} $\Rightarrow$ \texttt{lane\_index: 3}.
            \item E.g., ``moving forward in the leftmost lane'' with \texttt{lane\_number: 2} $\Rightarrow$ \texttt{lane\_index: 1}.
            \item E.g., ``parked by the curbside in the opposite direction'' with \texttt{lane\_number: 2} $\Rightarrow$ \texttt{lane\_index: -2}.
        \end{itemize}
    \end{itemize}
\end{enumerate}

\vspace{2ex}
\begin{enumerate}[label={Step \arabic*.}, start=10]
    \item
    Based on the information extracted in Steps 1--9, generate a structured YAML block. Wrap this with \texttt{{\textless}YAML{\textgreater}...{\textless}/YAML{\textgreater}} as shown below.
\end{enumerate}

\begin{tcolorbox}[
    breakable,
    enhanced,
    boxrule=0.8pt,
    colback=gray!5,
    colframe=gray!60,
    left=6pt, right=6pt, top=6pt, bottom=6pt,
]
\ttfamily\small
\begin{Verbatim}[breaklines=true, commandchars=\\\{\}]
<YAML>
environment:
    weather: clear  # one of: clear, cloudy, foggy, rainy, snowy, unspecified
    time: daytime   # one of: daytime, nighttime, unspecified
road_network:
    road_type: straight road  # one of: straight road, curved road,
                              #   intersection, t-intersection, roundabout,
                              #   unspecified
    road_context: urban  # one of: urban, residential, highway, country road,
                         #   dirt road, hill road, unspecified
    traffic_sign: none   # one of: stop sign, speed limit sign, yield sign,
                         #   yield to pedestrian sign, school zone, none,
                         #   unspecified
    traffic_light: none  # one of: green, red, yellow, broken, none, unspecified
    lane_number: 1       # integer or unspecified
    one-way_or_two-way: two-way  # one of: one-way, two-way, unspecified
actors:
    ego_vehicle:
        type: none           # always none for ego_vehicle
        behavior: move forward  # e.g.: move forward, stopped, turn left,
                                #   turn right, merge left, merge right,
                                #   unspecified
        speed: 30 km/h       # integer in km/h, or unspecified
        position_target: unspecified
        position_relation: unspecified  # one of: front, left, right, behind,
                                        #   left front, left behind, right front,
                                        #   right behind, unspecified
        lane_index: 1        # integer or unspecified
    other_actor_1:
        type: sedan          # e.g.: sedan, truck, tram, van, car, SUV, bus,
                             #   pedestrian, bicycle
        number: 1            # integer or unspecified
        behavior: move forward
        speed: 30 km/h       # integer in km/h, or unspecified
        position_target: ego_vehicle  # NOTE: always ego_vehicle for all
                                      #   actors other than ego_vehicle
        position_relation: front
        lane_index: 1
    ... (other actors if any) ...
</YAML>
\end{Verbatim}
\end{tcolorbox}

\vspace{2ex}
\textbf{\color{slategray}IN CONTEXT LEARNING EXAMPLE}

\textbf{Example input 1:}

The traffic scenario takes place on a two-lane, two-way urban street with no special lanes and no visible traffic signs or signals.
The environment is during daytime with cloudy weather. The roadway is lined with commercial buildings and sidewalks populated by pedestrians and cyclists.
Traffic is moderate, with vehicles moving smoothly and no apparent congestion, accidents, or hazards. No roadworks or lane closures are present. No emergency vehicles are observed.

\textbf{Actors}

The ego vehicle travels at about 28 km/h.
Ahead of it, a yellow sedan moves forward at approximately 30 km/h, followed by a black SUV traveling at around 32 km/h.
In the opposite lane, around four cars drive toward the ego at roughly 35 km/h.

\textbf{Example output 1:}
\begin{tcolorbox}[
    breakable,
    enhanced,
    width=0.65\textwidth,
    boxrule=0.8pt,
    colback=gray!5,
    colframe=gray!60,
    left=6pt, right=6pt, top=6pt, bottom=6pt,
]
\ttfamily\small
\begin{Verbatim}[breaklines=true, commandchars=\\\{\}]
<YAML>
environment:
    time: daytime
    weather: cloudy
road_network:
    lane_number: 1
    one-way_or_two-way: two-way
    road_type: straight road
    traffic_light: none
    traffic_sign: none
    road_context: urban
actors:
    ego_vehicle:
        type: none
        behavior: move forward
        lane_index: 1
        position_relation: unspecified
        position_target: unspecified
        speed: 28 km/h
    vehicle_in_front_1:
        type: sedan
        number: 1
        behavior: move forward
        lane_index: 1
        position_relation: front
        position_target: ego_vehicle
        speed: 30 km/h
    vehicle_in_front_2:
        type: SUV
        number: 1
        behavior: move forward
        lane_index: 1
        position_relation: front
        position_target: ego_vehicle
        speed: 32 km/h
    vehicles_opposite_direction:
        type: car
        number: 4
        behavior: move forward
        lane_index: -1
        position_relation: front
        position_target: ego_vehicle
        speed: 35 km/h
</YAML>
\end{Verbatim}
\end{tcolorbox}

\textbf{Example input 2:}

The traffic scenario takes place on a rural, two-way curved road within mountainous terrain.
The road has one lane in each direction, two lanes in total.
No traffic signs or lights are visible in the scene.
The environment features mostly clear daytime weather, providing good visibility of the surrounding rugged landscape.
Road work is in progress, as shown by a worker standing ahead in the ego vehicle's direction.
Traffic is light, with no congestion, accidents, or emergency vehicles present.
The roadside is characterized by rocky slopes and sparse trees, with wide open areas extending along both sides of the roadway.

\textbf{Actors}

The ego vehicle is moving forward at about 45 km/h. In the opposite lane, two construction trucks are traveling extremely slowly at around 3 km/h. Ahead of the ego vehicle, a road worker is standing still in the lane, remaining stationary at 0 km/h.

\textbf{Example output 2:}
\begin{tcolorbox}[
    breakable,
    enhanced,
    width=0.65\textwidth,
    boxrule=0.8pt,
    colback=gray!5,
    colframe=gray!60,
    left=6pt, right=6pt, top=6pt, bottom=6pt,
]
\ttfamily\small
\begin{Verbatim}[breaklines=true, commandchars=\\\{\}]
<YAML>
environment:
    time: daytime
    weather: clear
road_network:
    lane_number: 1
    one-way_or_two-way: two-way
    road_type: curved road
    traffic_light: none
    traffic_sign: none
    road_context: country road
actors:
    ego_vehicle:
        type: none
        behavior: move forward
        lane_index: 1
        position_relation: unspecified
        position_target: unspecified
        speed: 45 km/h
    construction_vehicles:
        type: construction vehicle
        number: 2
        behavior: move forward
        lane_index: -1
        position_relation: front
        position_target: ego_vehicle
        speed: 3 km/h
    traffic_worker:
        type: construction worker
        number: 1
        behavior: stopped
        lane_index: 1
        position_relation: front
        position_target: ego_vehicle
        speed: 0 km/h
</YAML>
\end{Verbatim}
\end{tcolorbox}

\textbf{Input natural language description:}

[Insert the input natural language description here.]
\label{fig:supp-prompt-description-to-dsl}
\end{tcolorbox}

%% file: supp/supp_tab_dsl_validation.tex
\begin{table}[htb]
\centering
\caption{\textbf{Driving model testing \& fine-tuning experiments.}
This table presents the average test results on {\numads} distinct autonomous driving models using scenarios generated with and without the DSL component. 
CR: collision rate, RR: frequency of running red lights, SS: frequency of running stop signs, OR: average distance driven out of road, RF: route following stability, Comp: average percentage of route completion, TS: average time spent to complete the route, ACC: average acceleration, YV: average yaw velocity, LI: frequency of lane invasion, OS: overall score.
$\uparrow$/$\downarrow$: higher/lower values indicate more effective in testing the driving models.
\textbf{Values in bold} indicate the best performance; 
}
\label{supp:tab-dsl-validation}

\centering
\begin{subtable}[t]{0.78\linewidth}
\centering
\caption{Testing driving models.}
\resizebox{\linewidth}{!}{%

\begin{tabular}{l|cccc|ccc|ccc|c}
\toprule
\multicolumn{1}{c|}{\multirow{2}{*}{\textbf{Method}}} & \multicolumn{4}{c|}{\textbf{Safety Level}} & \multicolumn{3}{c|}{\textbf{Functionality Level}} & \multicolumn{3}{c|}{\textbf{Etiquette Level}} & \multirow{2}{*}{\textbf{OS $\downarrow$}} \\
\multicolumn{1}{c|}{} & CR $\uparrow$ & RR $\uparrow$ & SS $\uparrow$ & OR $\uparrow$ & RF $\downarrow$ & Comp $\downarrow$ & TS $\uparrow$ & ACC $\uparrow$ & YV $\uparrow$ & LI $\uparrow$ &  \\ \midrule
w/o DSL Validation & 0.873 & 0.286 & 0.219 & 0.038 & 0.718 & \textbf{0.496} & 0.381 & 0.659 & 0.488 & 0.219 & 0.408 \\
with DSL Validation & \textbf{0.925} & \textbf{0.461} & \textbf{0.273} & \textbf{0.041} & \textbf{0.699} & 0.527 & \textbf{0.391} & \textbf{0.793} & \textbf{0.541} & \textbf{0.357} & \textbf{0.319} \\ \bottomrule
\end{tabular}%

}
\end{subtable}
\vfill
\vspace{1em}
\centering
\begin{subtable}[t]{0.78\linewidth}
\centering
\caption{Fine-tuning driving models.}

\resizebox{\linewidth}{!}{%
\begin{tabular}{l|cccc|ccc|ccc|c}
\toprule
\multicolumn{1}{c|}{\multirow{2}{*}{\textbf{Method}}} & \multicolumn{4}{c|}{\textbf{Safety Level}} & \multicolumn{3}{c|}{\textbf{Functionality Level}} & \multicolumn{3}{c|}{\textbf{Etiquette Level}} & \multirow{2}{*}{\textbf{OS $\uparrow$}} \\
\multicolumn{1}{c|}{} & CR $\downarrow$ & RR $\downarrow$ & SS $\downarrow$ & OR $\downarrow$ & RF $\uparrow$ & Comp $\uparrow$ & TS $\downarrow$ & ACC $\downarrow$ & YV $\downarrow$ & LI $\downarrow$ &  \\ \midrule
w/o DSL Validation & 0.745 & \textbf{0.318} & 0.010 & 0.000 & 0.749 & 0.533 & 0.312 & 0.574 & 0.501 & 0.418 & 0.369 \\
with DSL Validation & \textbf{0.691} & 0.366 & \textbf{0.000} & 0.000 & \textbf{0.814} & \textbf{0.593} & \textbf{0.211} & \textbf{0.553} & \textbf{0.496} & \textbf{0.381} & \textbf{0.402} \\ \bottomrule
\end{tabular}%

}
\end{subtable}

\end{table}

%% file: supp/tab_datasets.tex
\begin{table}[htb]
\centering
\caption{Summary statistics of the six LLM alignment datasets across diverse geographical locations.}
\label{supp:tab:datasets}

\resizebox{0.38\textwidth}{!}{%

\begin{tabular}{lccc}
\toprule
\textbf{Location} & \textbf{\# Videos} & \textbf{Avg. \# Actors} \\ \midrule
Los Angeles       & 82 & 13.8 \\
New York City     & 84 & 13.3 \\
Yellowstone       & 14 & \phantom{0}2.5  \\
Yosemite          & 20 & \phantom{0}2.3  \\
Small Towns in PA & 15 & \phantom{0}3.2  \\
Switzerland       & 46 & 11.1 \\ \bottomrule
\end{tabular}%

}

\end{table}

%% file: supp/supp_fig_dataset_diff.tex
\begin{figure}[htb]
  \centering
  \begin{subfigure}{0.8\linewidth}
    \centering
      \begin{subfigure}{0.49\linewidth}
        \centering
        \includegraphics[width=0.95\linewidth]{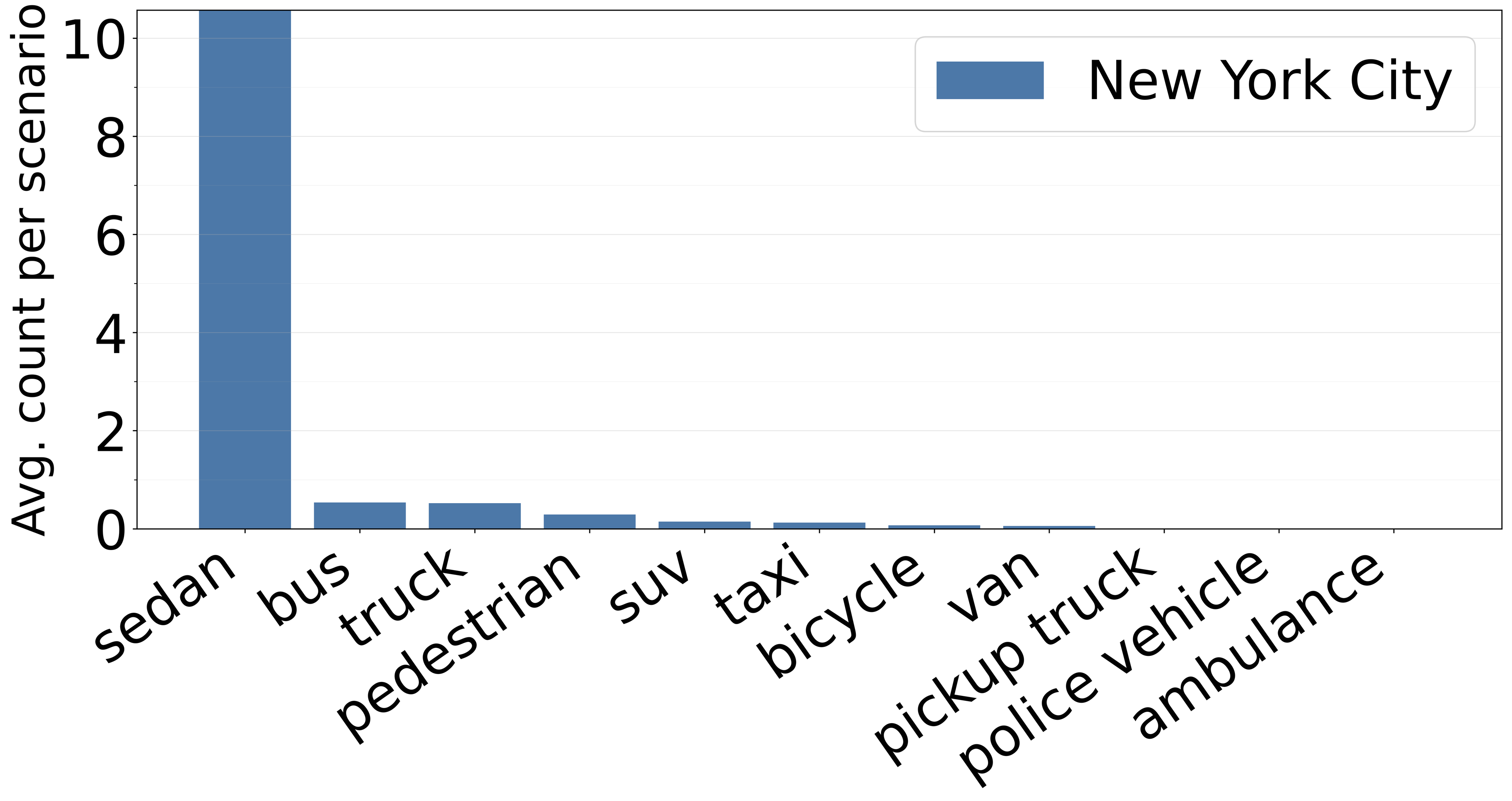}
      \end{subfigure}
      \hfill
      \begin{subfigure}{0.49\linewidth}
        \centering
        \includegraphics[width=0.95\linewidth]{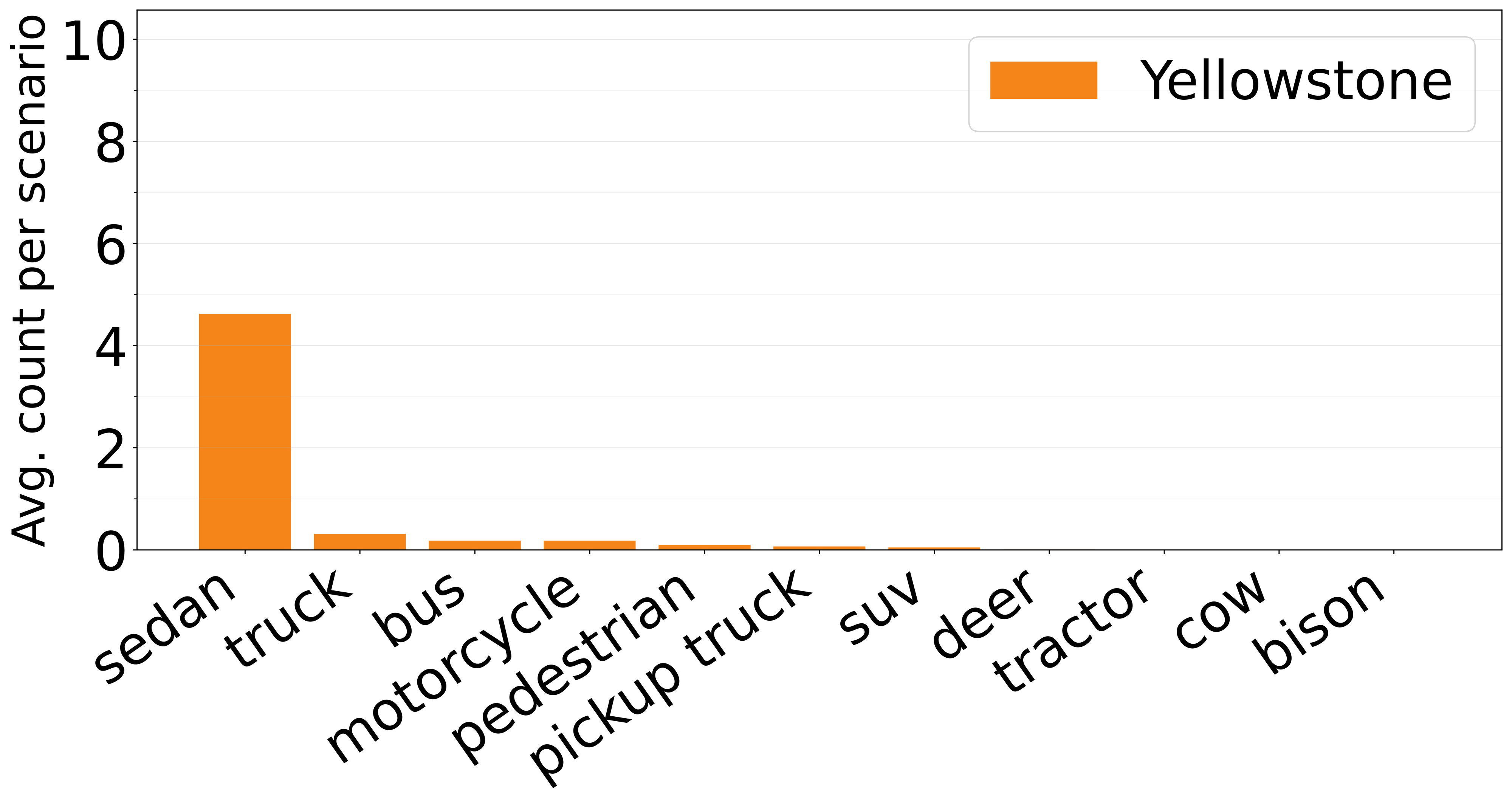}
      \end{subfigure}
    \caption{
    Actor type distributions.
    }
    \label{rebut:fig-hist-actor-type-behavior}
  \end{subfigure}
  \begin{subfigure}{0.8\linewidth}
    \centering
      \begin{subfigure}{0.49\linewidth}
        \centering
        \includegraphics[width=\linewidth]{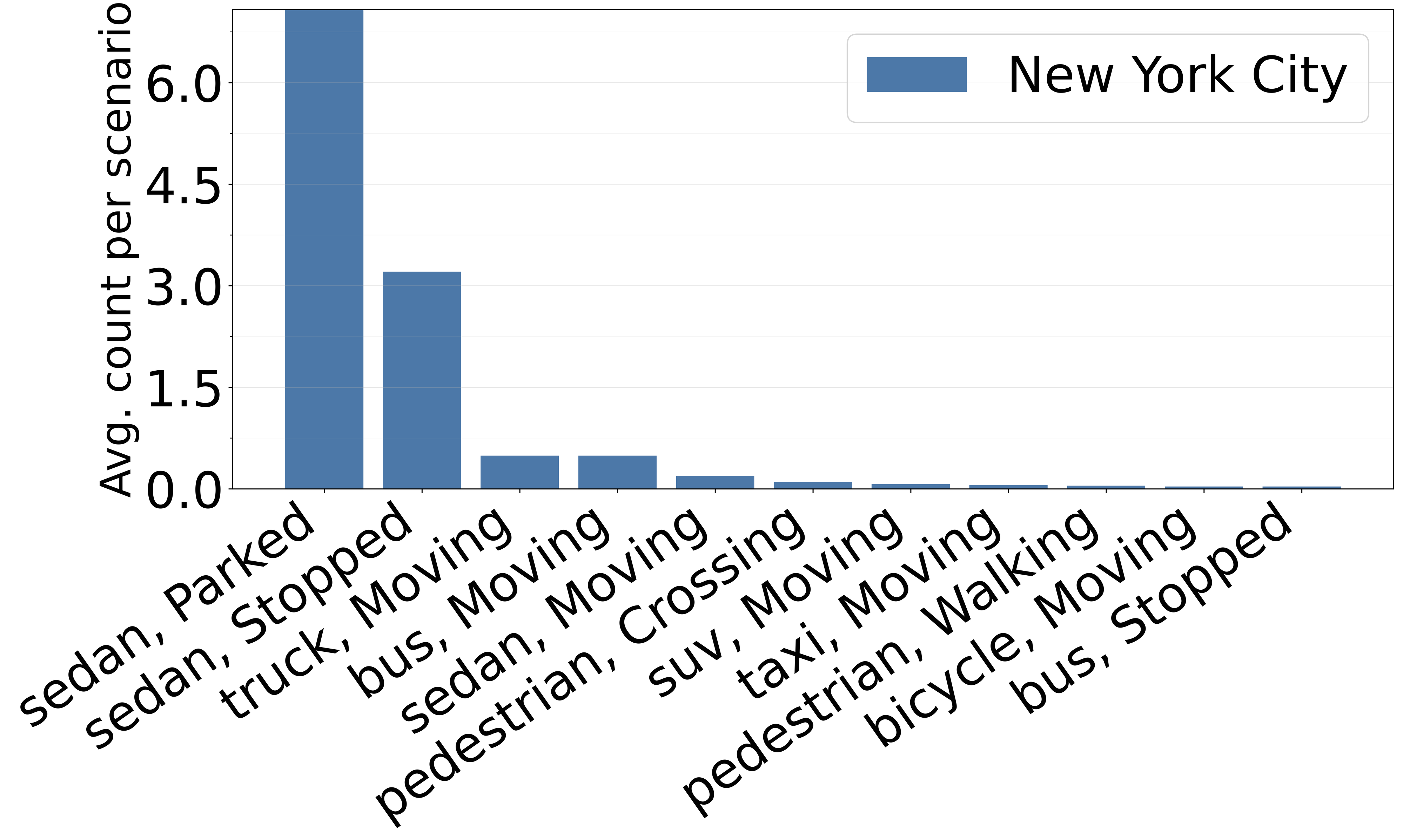}
      \end{subfigure}      
      \hfill
      \begin{subfigure}{0.49\linewidth}
        \centering
        \includegraphics[width=\linewidth]{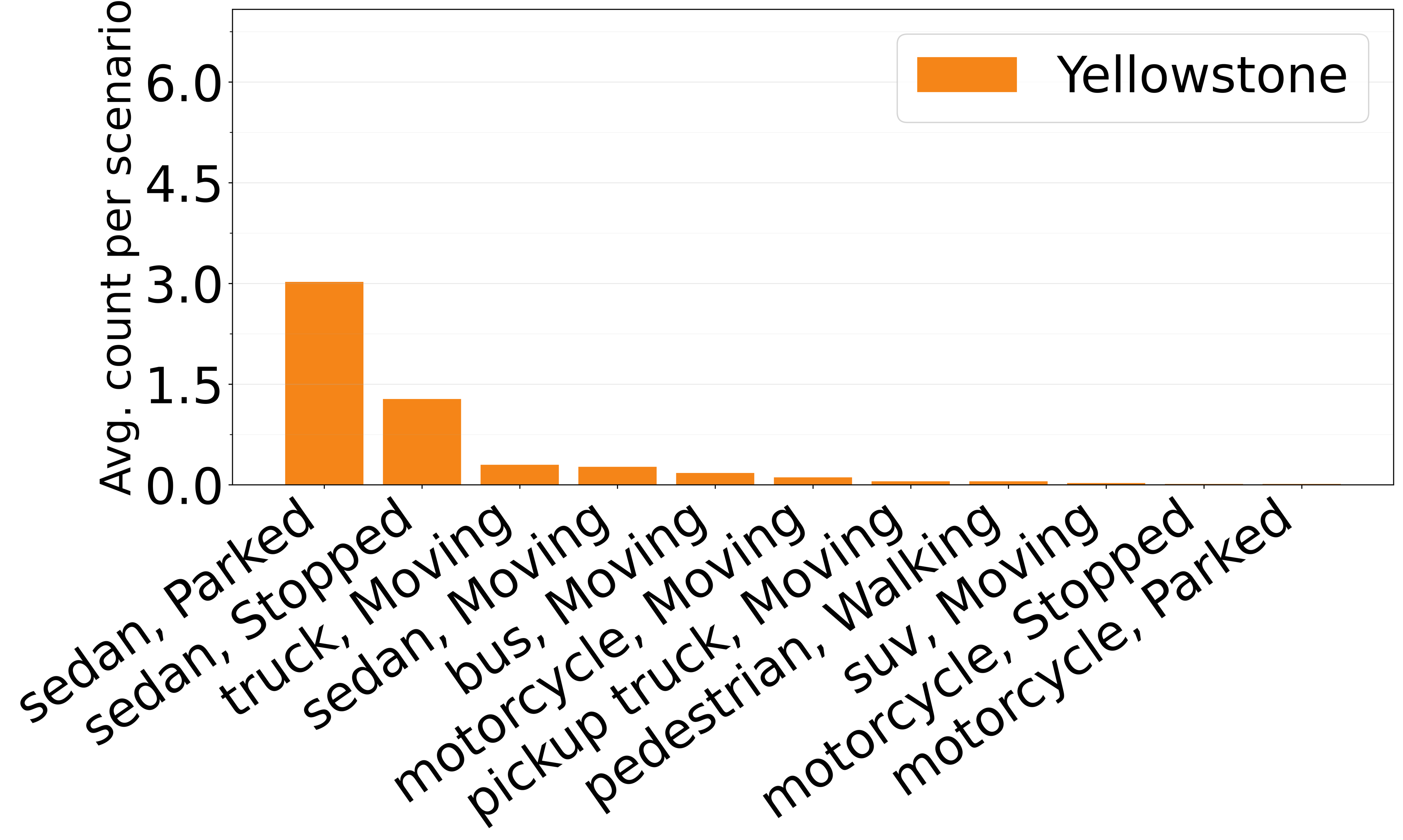}
      \end{subfigure}
    \caption{
    Actor type-behavior distributions.
    }
  \end{subfigure}
\caption{Behavioral differences across geographic regions.}
\label{supp:fig-dataset-diff}
\end{figure}

%% file: supp/supp_prompt_ours_aligned.tex
\begin{tcolorbox}[
    breakable,
    enhanced jigsaw, 
    width=\textwidth,
    colframe=black,
    colback=white,
    boxrule=1pt,
    arc=2mm,
    title={\textbf{Prompt for Aligned LLMs to Generate Traffic Scenarios}},
    fonttitle=\bfseries\large,
    coltitle=white,
    colbacktitle=black,
    center title,
    bottom=1mm,
    top=1mm,
    boxsep=3mm,
    before skip=10pt,
    after skip=15pt,
]

\textbf{\color{slategray}SYSTEM}

You are a traffic domain expert. 

\vspace{6pt}
\textbf{\color{slategray}TASK}

Design a traffic scenario in real life and provide a detailed description including the environment (weather, time of day), road network (road type, road context, lane number, one-way or two-way). 
Then design the position, behavior and speed of the ego vehicle and details of other actors, specifying actor type, behavior, speed, position, lane index, and any interactions with the ego vehicle. 
Wrap the generated textual description in \textless TEXT{\textgreater}...\textless /TEXT{\textgreater} tags.
\end{tcolorbox}

%% file: supp/supp_prompt_LLMbaselines.tex
\begin{tcolorbox}[
    breakable,
    enhanced jigsaw, 
    width=\textwidth,
    colframe=black,
    colback=white,
    boxrule=1pt,
    arc=2mm,
    title={\textbf{Prompt for Unaligned LLM baselines to Generate Traffic Scenarios}},
    fonttitle=\bfseries\large,
    coltitle=white,
    colbacktitle=black,
    center title,
    bottom=1mm,
    top=1mm,
    boxsep=3mm,
    before skip=10pt,
    after skip=15pt,
]

\textbf{\color{slategray}SYSTEM}

You are a traffic domain expert. Your task is to design a traffic scenario for testing autonomous driving systems.

\vspace{6pt}
\textbf{\color{slategray}TASK}

Design a traffic scenario take place in \textbf{[Insert the region name here]}.
Approach this task step-by-step, take your time, and don’t skip steps.

Design the basic configurations of the traffic scenario:
\begin{enumerate} [label={Step \arabic*.}]
    \item
    weather: one of [clear, cloudy, foggy, rainy, snowy], or describe in your own words
    \item
    time: one of [daytime, nighttime], or describe in your own words
    \item
    road type: one of [straight road, intersection, t-intersection, roundabout]
    \item
    one-way or two-way: one of [one-way, two-way]
    \item
    special lane: if there are any special lanes (e.g., bike lane, tram track, bus lane)
    \item
    lane number: use an integer. 
    \item
    traffic sign: one of [stop sign, speed limit sign, yield sign, yield to pedestrian sign, school zone], or state   ``none''
    \item
    traffic light: one of [green, red, yellow, broken], or state ``none''
    \item
    road context: one of [urban, residential, highway, country road, dirt road, hill road], or describe in your own words
    \item
    Design additional and detailed information of the traffic scenario:
    \begin{itemize}
        \item Whether the traffic is heavy or light; Is there a traffic jam?
        \item Whether there are any traffic accidents or other road hazards
        \item Whether there are any road works or lane closures
        \item Whether there are any emergency vehicles
        \item Describe the environment of the roadside, including the buildings, trees, and other objects.
    \end{itemize}

    \item
    Consider any other information that you think is important.
    
    \item
    actors: For each detected actor (e.g., car, bus, pedestrian, bicycle), record the following attributes:
    \begin{itemize}
        \item type: The classification of the actor (e.g., sedan, SUV, truck, pedestrian, cyclist)
        \item current behavior: The actor's present action or intent (e.g., move forward, turn left, stop, yield)
        \item speed: The current speed of the actor, measured in kilometers per hour (km/h)
        \item position: The actor's location, which should be described in two ways: (1) relative position to the ego vehicle using position target and position relation, and (2) the absolute position in the road network using lane index
        \item position target and position relation: Relative position to the ego vehicle. Specify the actor's position concerning the ego vehicle.
        \begin{itemize}
            \item For example, if another actor is ``directly ahead of the ego vehicle in the same lane'', position target is ego vehicle; position relation is front.
            \item Another example, if another actor is ``to the left rear of the ego vehicle in an adjacent lane'', position target is ego vehicle; position relation is left behind.
        \end{itemize}
        
        \item lane index: an integer stating the lane index within the road network. The lane index of the leftmost lane in the direction of the ego vehicle is 1. The lane index of the leftmost lane in the opposite direction of the ego vehicle is -1.
        \begin{itemize}
            \item For example, if an actor is ``parked in the rightmost lane'', and the lane number is 3, then the lane index is 3.
            \item Another example, if an actor is ``moving forward in the leftmost lane'' and the lane number is 2, then the lane index is 1.
        \end{itemize}
    \end{itemize}
\end{enumerate}

Generate a concise, technical description of the traffic scenario that:
\begin{itemize}
    \item Use precise traffic engineering terminology.
    \item Use languages to describe a traffic scenario rather than a image.
    \item Use active voice and present tense.
    \item Avoid subjective observations.
    \item Use assertive language (``is'' / ``are'' instead of ``appears'' / ``seems'' / ``possibly'').
    \item Avoid image-related terms (e.g., ``shows'', ``displays'', ``depicts'').
    \item Use precise numbers (e.g., ``10 cars'' instead of ``many cars'' / ``multiple cars'').
    \item Maintain all key information about the road network, the environment, and actors.
    \item Describe the road network first, then the environment, then the actors. 
    \item For the road network, first describe the road context and the road type, one-way / two-way, and any special lanes (e.g., bike lane, tram track). Then describe the lane number, the traffic sign, and the traffic light.
    \item For the environment, describe the weather and time of day.
    \item Describe the additional and detailed information observed in step 10 and step 11.
    \item At the end, add the descriptions of the actors. For the actors, describe the type and count of each type of actor. Remember to include the precise number (in digits) of actors.
    \item Treat the actor section differently and start this section with a separate title **Actors** in a new line.
\end{itemize}

Start your generated description with \textless TEXT{\textgreater} and end with \textless /TEXT{\textgreater}

\vspace{6pt}
\textbf{\color{slategray}OUTPUT FORMAT}

\textless ANALYSIS{\textgreater}

[Your chain of thoughts following the step-by-step instructions provided in step 1 to step 12]

\textless /ANALYSIS{\textgreater}

\textless TEXT{\textgreater}

[Insert paragraph description here]

**Actors**

[Insert paragraph describing the actors here]

\textless /TEXT{\textgreater}
\end{tcolorbox}